\documentclass{article}
\usepackage{ijcai21}
\usepackage{times}
\usepackage{soul}
\usepackage{url}
\usepackage[hidelinks]{hyperref}
\usepackage[utf8]{inputenc}
\usepackage[small]{caption}
\usepackage{graphicx}
\usepackage{amsmath}
\usepackage{amsthm}
\usepackage{booktabs}
\usepackage{algorithm}
\usepackage{algorithmic}
\usepackage{amssymb}
\usepackage{multirow}
\usepackage{hyperref}
\usepackage{subfigure}
\usepackage{tikz, pgfplots}
\usetikzlibrary{intersections}
\usepgfplotslibrary{fillbetween}
\usepackage{wrapfig}
\usepackage{url}
\usepackage{sidecap}
\usepackage{footnote}
\usepackage{threeparttable}
\usepackage{lipsum}
\usepackage{bm}
\usepackage{float}

\newtheorem{proposition}{Proposition}

\newtheorem{lamma}{Lamma}

\title{Hindsight Trust Region Policy Optimization}

\author{
Hanbo Zhang$^1$\and
Site Bai$^1$\and
Xuguang Lan$^{1}$\footnote{Corresponding Author}\and
David Hsu$^2$\And
Nanning Zheng$^1$\\
\affiliations
$^1$Xi'an Jiaotong University\\
$^2$National University of Singapore\\
\emails
\{zhanghanbo163, best99317\}@stu.xjtu.edu.cn,
xglan@xjtu.edu.cn,
dyhsu@comp.nus.edu.sg,
nnzheng@xjtu.edu.cn
}

\begin{document}

\maketitle

\begin{abstract}
Reinforcement Learning (RL) with sparse rewards is a major challenge. We propose Hindsight Trust Region Policy Optimization (HTRPO), a new RL algorithm that extends the highly successful TRPO algorithm with hindsight to tackle the challenge of sparse rewards. Hindsight refers to the algorithm's ability to learn from information across goals, including past goals not intended for the current task. We derive the hindsight form of TRPO, together with QKL, a quadratic approximation to the KL divergence constraint on the trust region. QKL reduces variance in KL divergence estimation and improves stability in policy updates.  We show that HTRPO has similar convergence property as TRPO. We also present Hindsight Goal Filtering (HGF), which further improves the learning performance for suitable tasks. HTRPO has been evaluated on various sparse-reward tasks, including Atari games and simulated robot control. Results show that HTRPO consistently outperforms TRPO, as well as HPG, a state-of-the-art policy gradient algorithm for RL with sparse rewards.
\end{abstract}

\section{Introduction}

Reinforcement Learning (RL) has been widely investigated to solve problems from complex strategic games \cite{mnih2015human} to precise robotic control \cite{deisenroth2013survey}. However, current successful practice of RL in robotics relies heavily on careful and arduous reward shaping\cite{ng1999policy,grzes2017reward}. \textbf{Sparse reward}, in which the agent is rewarded only upon reaching the desired goal, obviates designing a delicate reward mechanism. It also guarantees that the agent focuses on the intended task itself without any deviation. However, sparse reward diminishes the chance for policy to converge, especially in the initial random exploration stage, since the agent can hardly get positive feedbacks.

Recently, several works have been devoted to sparse-reward RL. \cite{andrychowicz2017hindsight} proposes Hindsight Experience Replay(HER), which trains the agent with hindsight goals generated from the achieved states through the historical interactions. Such hindsight experience substantially alleviates exploration problem caused by sparse-reward settings. \cite{rauber2018hindsight} proposes Hindsight Policy Gradient(HPG). It introduces hindsight to policy gradient, resulting in an advanced algorithm for RL with sparse reward. However, for HPG, there remain several drawbacks hindering its application in more cases. Firstly, as an extension to ``vanilla'' policy gradient, its performance level and sample efficiency remain limited. Secondly, it inherits the intrinsic high variance of PG methods, and the combination with hindsight data further exacerbates the learning stability.

In this paper, we propose \textbf{Hindsight Trust Region Policy Optimization (HTRPO)}, a hindsight form of TRPO~\cite{schulman2015high}, which is an advanced RL algorithm with approximately monotonic policy improvements.  We prove that HTRPO theoretically inherits the convergence property of TRPO, and significantly reduces the variance of policy improvement by introducing Quadratic KL divergence Estimation (QKL) approach. Moreover, to select hindsight goals that better assist the agent to reach the original goals, we design a Hindsight Goal Filtering mechanism.

We demonstrate that in a wide variety of sparse-reward tasks including benchmark toy tasks, image-input Atari games and discrete and continuous robotic control, HTRPO can consistently outperform TRPO and HPG in the aspect of performance and sample efficiency with commendable learning stability. We also provide a comprehensive comparison with HER, showing that HTRPO achieves much better performance in 6 out of 7 benchmarks.
Besides, we also conduct ablation studies to show that Quadratic KL divergence Estimation can effectively lower the variance and constrain the divergence while Hindsight Goal Filtering brings the performance to a higher level especially in more challenging tasks.

\section{Preliminaries}

\paragraph{RL Formulation and Notation.} \
Consider the standard infinite-horizon reinforcement learning formulation which can be defined by tuple $(\mathcal{S}, \mathcal{A}, \mathcal{\pi}, \rho_0, r, \gamma)$. $ \mathcal{S} $ and $ \mathcal{A} $ denote the set of states and actions respectively. $\pi: \mathcal{S} \to \mathcal{P} (\mathcal{A}) $ is a policy mapping states to a distribution over actions. $\rho_0$ is the distribution of the initial state $s_0$. Reward function $r: \mathcal{S} \to \mathbb{R}$ defines the reward obtained from the environment and $\gamma \in (0, 1)$ is a discount factor. In this paper, the policy is a differentiable function regarding parameter $\theta$. We follow the standard formalism of state-action value function $Q(s,a)$, state value function $V(s)$ and advantage function $A(s,a)$ in \cite{sutton2018reinforcement}. We also adopt the definition of $\gamma$-discounted state visitation distribution as $\rho_{\theta}(s) = (1-\gamma) \sum_{t=0}^{\infty} \gamma^t P(s_t = s)$ \cite{ho2016model}. Correspondingly, $\gamma$-discounted state-action visitation distribution \cite{ho2016model}, also known as occupancy measure \cite{ho2016generative}, is defined as $\rho_{\theta}(s,a)=\rho_{\theta}(s)\times \pi_{\theta}(a|s)$.

\paragraph{Trust Region Policy Optimization(TRPO).} \
TRPO \cite{schulman2015trust} is an iterative trust region method that effectively optimizes policy by maximizing the per-iteration policy improvement. \ The optimization problem proposed in TRPO can be formalized as follows:
\begin{align}
\max_{\theta} \ \mathop{\mathbb{E}}_{s, a \sim \rho_{\tilde{\theta}}(s,a)} \left [ \frac{\pi_{\theta}(a|s)}{\pi_{\tilde{\theta}}(a|s)} A_{\tilde{\theta}}(s, a) \right ] \label{eq1.2}
\end{align}
\vspace{-12pt}
\begin{align}
s.t. \ \mathop{\mathbb{E}}_{s \sim \rho_{\tilde{\theta}}(s)} \left[ D_{KL}(\pi_{\tilde{\theta}}(a|s)||\pi_{\theta}(a|s)) \right] \leq \epsilon \label{eq1.1}
\end{align}
in which $\rho_{\tilde{\theta}}(s) = \sum_{t=0}^{\infty} \gamma^t P(s_t = s)$. $\theta$ denotes the parameter of the new policy while $\tilde{\theta}$ is that of the old one.
\paragraph{Hindsight Policy Gradient(HPG).} \
HPG~\cite{rauber2018hindsight} combines the idea of hindsight~\cite{andrychowicz2017hindsight} with policy gradient methods. Though goal-conditioned reinforcement learning has been explored for a long time and actively investigated in recent works \cite{peters2008reinforcement,schaul2015universal,veeriah2018many}, HPG firstly extends the idea of hindsight to goal-conditioned policy gradient and  shows that the policy gradient can be computed in expectation over all goals. The goal-conditioned policy gradient is derived as follows:
\begin{align}
\nabla_{\theta} \eta(\pi_\theta) = \mathop{\mathbb{E}}_{g, \tau} \left [ \sum_{t=0}^{T-1} \nabla_{\theta} \log \pi_{\theta}(a_t | s_t, g) A_{\theta}(s_t, a_t, g) \right ] \label{eq1.3}
\end{align}
where $\tau \sim p_{\theta}(\tau | g)$. Then, by applying hindsight formulation, it rewrites goal-conditioned policy gradient with trajectories conditioned on achieved goal $g'$ using importance sampling to solve sparse-reward problems efficiently.


\section{Hindsight Trust Region Policy Optimization}

In this section, we firstly introduce Quadratic KL divergence Estimation (QKL) method, which efficiently reduces the variance of KL estimation in TRPO and results in higher learning stability. With QKL, we show that TRPO maintains the monotonically-converging property. After that, we derive the hindsight form of TRPO, called Hindsight Trust Region Policy Optimization algorithm, to tackle the severely off-policy hindsight data for better learning with sparse rewards. Specifically, the expected return and the KL divergence constraint are both modified to adapt to hindsight data with importance sampling. Benefiting from QKL, we can precisely estimate KL divergence using hindsight data while keeping the variance below a reasonable level. Intuitively, HTRPO utilizes hindsight data to estimate the objective and the constraint, and iteratively find out the local optimal policy to ensure the approximately monotonous policy improvements.

\subsection{TRPO with Quadratic KL Divergence} \label{cons}


In TRPO, the KL divergence expectation under $\rho_{\tilde{\theta}}(s)$ is estimated by averaging values of KL divergence conditioned on collected states. 
However, this method is no longer valid if KL divergence cannot be analytically computed (e.g. Gaussian Mixture Model) or the state distribution changes (e.g. using hindsight data instead of the collected ones). 
To solve this problem, we firstly transform the KL divergence to an expectation under occupancy measure $\rho_{\tilde{\theta}}(s,a)=\rho_{\tilde{\theta}}(s)\times \pi_{\tilde{\theta}}(a|s)$. 
It can be estimated using the collected state-action pairs $(s,a)$, no longer depending on the analytical form of KL divergence. 
Also, such formulation is convenient for correcting changed distribution over state and action by importance sampling, which will be discussed in section \ref{obj}. 
However, it will increase the estimation variance, causing instability of training. 
Therefore, by making use of another $f$-divergence, we propose QKL to approximate KL divergence for variance reduction, and both theoretically and practically, we prove the effectiveness of such an approximation.

Given two policies $\pi_{\tilde{\theta}}(a|s)$ and $\pi_{\theta}(a|s)$, the KL-divergence over state $s$ can be converted to a logarithmic form:
\begin{equation} \nonumber \label{eq3.1}
\resizebox{1\linewidth}{!}{$
    \displaystyle
    D_{KL}(\pi_{\tilde{\theta}}(a|s)|| \pi_{\theta}(a|s)) = \mathop{\mathbb{E}}_{a \sim \pi_{\tilde{\theta}}(a|s)} \left[ \log \pi_{\tilde{\theta}}(a|s) - \log \pi_{\theta}(a|s) \right] 
$}
\end{equation}
However, simply expanding the KL-divergence into logarithmic form still leaves several problems unhandled. Firstly, such formulation causes excessively high estimation variance.
Secondly, such estimation of KL-divergence is of possible negativity. 
To overcome these two drawbacks, we propose Quadratic KL Divergence Estimation in Proposition \ref{propos1} and prove that such approximation will reduce the estimation variance in Proposition \ref{propos2} (detailed proof can be found in Appendix \ref{ap:ap5} and \ref{ap:ap6}):

\begin{proposition}\label{propos1}
(Quadratic KL Divergence Estimation). For policy $\pi_{\tilde{\theta}}(a|s)$ and $\pi_{\theta}(a|s)$, and for $\eta = \pi_{\theta}(a|s) - \pi_{\tilde{\theta}}(a|s)$,
\begin{align}
\mathop{\mathbb{E}}_{a} &\left[ \log \pi_{\tilde{\theta}}(a|s) - \log \pi_{\theta}(a|s) \right ] \nonumber \\
&= \ \mathop{\mathbb{E}}_{a} \left[ \frac{1}{2} ( \log \pi_{\tilde{\theta}}(a|s) - \log \pi_{\theta}(a|s))^2 \right] + \mathop{\mathbb{E}}_{a} \left[ O( \eta^3 )  \right] \label{eq3.2}
\end{align}
where $a \sim \pi_{\tilde{\theta}}(a|s)$.
\end{proposition}

Proposition \ref{propos1} demonstrates that when $\theta$ and $\tilde{\theta}$ is of limited difference, the expectation of $\log \pi_{\tilde{\theta}}(a|s) - \log \pi_{\theta}(a|s)$ can be sufficiently estimated by the expectation of its square. In fact, $\mathop{\mathbb{E}}_{a \sim \pi_{\tilde{\theta}}(a|s)} \left[ \frac{1}{2} ( \log \pi_{\tilde{\theta}}(a|s) - \log \pi_{\theta}(a|s))^2 \right]$ is an $f$-divergence, where $f(x) = \frac{1}{2} x( \log x )^2$, which we call $D_{QKL}$ in this paper. Noticeably, though $f(x)$ is a convex function only when $x \in (\frac{1}{e}, \infty)$, and it indeed does not correspond to an $f$-divergence, in our practice, $\frac{\pi_{\tilde{\theta}}(a|s)}{\pi_{\theta}(a|s)} > \frac{1}{e}$ holds, hence we can define a convex function on $R^+$: $f(x)=\frac{1}{2} x( \log x )^2$ when $x \in (\frac{1}{e}, \infty)$ and $-x+\frac{2}{e}$ when $x \in (0, \frac{1}{e}]$, with an unused piece defined over $(0, \frac{1}{e}]$. 

\begin{proposition}\label{propos2}
(Variance of Constraint Function). For policy $\pi_{\tilde{\theta}}(a|s)$ and $\pi_{\theta}(a|s)$, let $\mathrm{Var}$ denote the variance of a variable. For any action $a \in \mathcal{A}$ and any state $s \in \mathcal{S}$, when $\log \pi_{\tilde{\theta}}(a|s) - \log \pi_{\theta}(a|s) \in \left[ -0.5 , 0.5 \right ]$, then
\begin{align}
\mathop{\mathrm{Var}}_{a \sim \pi_{\tilde{\theta}}(a|s)} \left[ \frac{( \log \pi_{\tilde{\theta}}(a|s) - \log \pi_{\theta}(a|s))^2}{2}  \right] \nonumber \\
\leq \mathop{\mathrm{Var}}_{a \sim \pi_{\tilde{\theta}}(a|s)} \left[ \log \pi_{\tilde{\theta}}(a|s) - \log \pi_{\theta}(a|s) \right]. \label{eq3.3}
\end{align}
\end{proposition}

Proposition \ref{propos2} illustrates that there is a decrease from the variance of $\log \pi_{\tilde{\theta}}(a|s) - \log \pi_{\theta}(a|s)$ to the variance of its square. In fact, the closer it is between $\tilde{\theta}$ and $\theta$, the more the variance decreases. Next, we will show that with the introduction of QKL, TRPO still maintains similar convergence property.

\begin{proposition}\label{propos3}
(Policy Improvement Guarantee) Given two policies $\pi_{\theta}$ and $\pi_{\tilde{\theta}}$, Let 
\begin{align}
s^*=\arg\max_s D_{TV}(\pi_{\tilde{\theta}}(a|s), \pi_{\theta}(a|s)) \nonumber
\end{align}
If $\frac{D_{KL}(\pi_{\tilde{\theta}}(a|s^*), \pi_{\theta}(a|s^*))} {D_{QKL}(\pi_{\tilde{\theta}}(a|s^*), \pi_{\theta}(a|s^*))} \leq \frac{2}{\ln 2}$, then 
\begin{align}
\eta(\pi_{\theta}) \geq  L_{\pi_{\tilde{\theta}}}(\pi_{\theta}) - CD_{QKL}^{max}(\pi_{\tilde{\theta}}(a|s), \pi_{\theta}(a|s))
\end{align}
where 
\begin{align}
L_{\pi_{\tilde{\theta}}}(\pi_{\theta}) = \eta(\pi_{\tilde{\theta}}) + E_{s\sim \pi_{\tilde{\theta}}(s), a\sim \pi_{\theta}(a|s)}[A_{\pi_{\tilde{\theta}}}(s, a)]
\end{align}
and $\eta(\pi_{\theta}) =E\left[\sum \gamma^t r_t\right]$ is the expected return, $C=\frac{4\beta\gamma}{(1-\gamma)^2}$, $\beta=max_{s,a}|A_{\pi_{\tilde{\theta}}} (s,a)|$, $D_{TV}(p,q) = \frac{1}{2}\sum_i\left|p_i-q_i\right|$.
\end{proposition}
The proof and detailed analysis are given in Appendix \ref{ap:ap-ca}. 
Intuitively, Proposition \ref{propos3} means that when two policies are not far from each other, the convergence property of TRPO also holds for the QKL constraint.
As a result, with Proposition \ref{propos3}, we can derive a new but similar monotonically-converging algorithm as in TRPO, given in Appendix \ref{ap:ap-pracqkltrpo}.
By taking a series of approximation as shown in Appendix \ref{ap:ap-pracqkltrpo}, the following policy optimization problem is derived, called QKL-TRPO.
\begin{align} \nonumber
\max_{\theta} \ \mathop{\mathbb{E}}_{s, a \sim \rho_{\tilde{\theta}}(s,a)} \left [ \frac{\pi_{\theta}(a|s)}{\pi_{\tilde{\theta}}(a|s)} A_{\tilde{\theta}}(s, a) \right ] \tag{\ref{eq1.2}}
\end{align}
\vspace{-10pt}
\begin{align}
s.t. \ \mathop{\mathbb{E}}_{s, a} \left[\frac{1}{2}( \log \pi_{\tilde{\theta}}(a|s) - \log \pi_{\theta}(a|s))^2 \right] \leq \epsilon \label{eq-qkltrpocons}
\end{align}
It is noteworthy that QKL-TRPO can be applied to policies which do not correspond to an analytic KL divergence (e.g. GMM policies). We also provide a simple analysis of QKL-TRPO compared with the original TRPO in Appendix \ref{ap:ap-qkltrpo}, which shows that QKL-TRPO is comparable with TRPO in a series of MuJoCo benchmarks.


\subsection{Hindsight Formulation of QKL-TRPO} \label{obj}

In this section, we derive the hindsight form of the QKL-TRPO, called Hindsight Trust Region Policy Optimization (HTRPO), to efficiently tackle severely off-policy hindsight experience and sparse-reward RL problems.

Starting from eq.\ref{eq1.2}, it can be written in the following variant form:
\begin{align}
L_{\tilde{\theta}}(\theta) &= \mathop{\mathbb{E}}_{\tau \sim p_{\tilde{\theta}}(\tau)} \left [ \sum_{t=0}^{\infty} \gamma^t \frac{\pi_{\theta}(a_{t}|s_{t})}{\pi_{\tilde{\theta}}(a_{t}|s_{t})} A_{\tilde{\theta} }(s_t, a_t) \right ] \label{eq2.0}
\end{align}
The derivation process of this variant form is shown explicitly in Appendix \ref{ap:ap1} and in \cite{schulman2015trust}. Given the expression above, similar to eq.\ref{eq1.3}, we consider the goal-conditioned objective function:
\begin{align}
L_{\tilde{\theta}}(\theta) &= \mathop{\mathbb{E}}_{g, \tau} \left [ \sum_{t=0}^{\infty} \gamma^t \frac{\pi_{\theta}(a_{t}|s_{t}, g)}{\pi_{\tilde{\theta}}(a_{t}|s_{t}, g)} A_{\tilde{\theta} }(s_t, a_t, g) \right ] \label{eq2.1}
\end{align}
where $\tau \sim p_{\tilde{\theta}}(\tau | g)$.
For the record, though it seems that eq.\ref{eq2.1} makes it possible for off-policy learning, it can be used as the objective only when policy ${\pi_\theta}$ is close to the old policy $\pi_{\tilde{\theta}}$, i.e. within the trust region. Using severely off-policy data like hindsight experience will make the learning process diverge. Therefore, importance sampling is integrated to correct the difference of the trajectory distribution caused by changing the goal. Based on eq.\ref{eq2.1}, the following Proposition gives out the hindsight objective function conditioned on some goal $g'$ with the distribution correction derived from importance sampling.

\begin{proposition}\label{ap:prop4}
(Hindsight Expected Return). For the original goal $g$ and hindsight goal $g'$, the object function of HTRPO $ L_{\tilde{\theta}}(\theta) $ is given by:
\begin{equation} \label{eq2.2}
\resizebox{\linewidth}{!}{$
    \displaystyle
    L_{\tilde{\theta}}(\theta)
= \mathop{\mathbb{E}}_{g', \tau} \left[ \sum_{t=0}^{\infty} \prod_{k=0}^{t} \frac{\pi_{\tilde{\theta}}(a_k | s_k, g')}{\pi_{\tilde{\theta}}(a_k | s_k, g)} \gamma^t \frac{\pi_\theta(a_t|s_t, g')}{\pi_{\tilde{\theta}}(a_t|s_t, g')}  A_{\tilde{\theta}}(s_t, a_t, g') \right ]
$}
\end{equation}
in which $\tau \sim p_{\theta}(\tau | g)$ and $\tau =  s_0, a_0, s_1, a_1, ... , s_t, a_t$. 
\end{proposition}
Appendix \ref{ap:ap:prop4sec} presents an explicit proof of how the hindsight-form objective function derives from eq.\ref{eq2.1}. In our practice, we introduce a baseline $V_\theta(s)$ for computing the advantage $A_\theta$. Though $A_\theta$ here can be estimated by combining per-decision return~\cite{precup2000eligibility}, due to its high variance, we adopt one-step TD method instead to get $A_\theta$, i.e., $A_\theta(s,a) = r(s,a)+\gamma V_\theta(s')-V_\theta(s)$.
Intuitively, eq.\ref{eq2.2} provides a way to compute the expected return in terms of the advantage with new-goal-conditioned hindsight experiences which are generated from interactions directed by old goals.

Next, we demonstrate that hindsight can also be introduced to the constraint function. The proof follows the methodology similar to that in Proposition \ref{ap:prop4}, and is deducted explicitly in Appendix \ref{ap:ap7}.

\begin{proposition}\label{propos5}
(HTRPO Constraint Function). For the original goal $g$ and hindsight goal $g'$, the constraint between policy $\pi_{\tilde{\theta}}(a|s)$ and policy $\pi_{\theta}(a|s)$ is given by:
\begin{align}
 \mathop{\mathbb{E}}_{g'} \left [ \mathop{\mathbb{E}}_{\tau \sim p_{\theta}(\tau | g)} \left [  \sum_{t = 0} ^\infty \prod_{k=0}^{t} \frac{\pi_{\tilde{\theta}}(a_k | s_k, g')}{\pi_{\tilde{\theta}}(a_k | s_k, g)} \gamma^t K_t \right] \right ] \leq \epsilon' \label{eq3.4}
\end{align}
in which $\epsilon' = \frac{\epsilon}{1 - \gamma}$, and $K_t=\frac{1}{2}(\log \pi_{\tilde{\theta}}(a_t|s_t, g') - \log \pi_{\theta}(a_t|s_t,g'))^2$.
\end{proposition}

Proposition \ref{propos5} implies the practicality of using hindsight data under condition $g'$ to estimate the KL expectation. From all illustration above, we give out the final form of the optimization problem for HTRPO:

\vspace{-10pt}

\begin{align}
\label{eq_htrpo_obj}
\vspace{-20pt}
\max_{\theta} \ \mathop{\mathbb{E}}_{g'} \left [  \mathop{\mathbb{E}}_{\tau \sim p_{\theta}(\tau | g)} \left[ \sum_{t=0}^{\infty} \prod_{k=0}^{t} \frac{\pi_{\tilde{\theta}}(a_k | s_k, g')}{\pi_{\tilde{\theta}}(a_k | s_k, g)}\gamma^t R_t  \right ] \right ]
\end{align}
\vspace{-10pt}
\begin{align}
\label{eq_htrpo_cons}
s.t. \ \mathop{\mathbb{E}}_{g'} \left [ \mathop{\mathbb{E}}_{\tau \sim p_{\theta}(\tau | g)} \left [ \sum_{t = 0} ^\infty \prod_{k=0}^{t} \frac{\pi_{\tilde{\theta}}(a_k | s_k, g')}{\pi_{\tilde{\theta}}(a_k | s_k, g)} \gamma^t K_t \right] \right ] \leq \epsilon'
\vspace{-10pt}
\end{align}
where $R_t=\frac{\pi_\theta(a_t|s_t, g')}{\pi_{\tilde{\theta}}(a_t|s_t, g')}  A_{\tilde{\theta}}(s_t, a_t, g')$ and $K_t=\frac{1}{2}(\log \pi_{\tilde{\theta}}(a_t|s_t, g') - \log \pi_{\theta}(a_t|s_t,g'))^2$. The solving process for HTRPO optimization problem is explicitly demonstrated in Appendix \ref{ap:ap8}.

\section{Hindsight Goal Filtering}
\label{sec_hgf}

\begin{figure}[t] 
 \center{\includegraphics[width=0.45\textwidth]{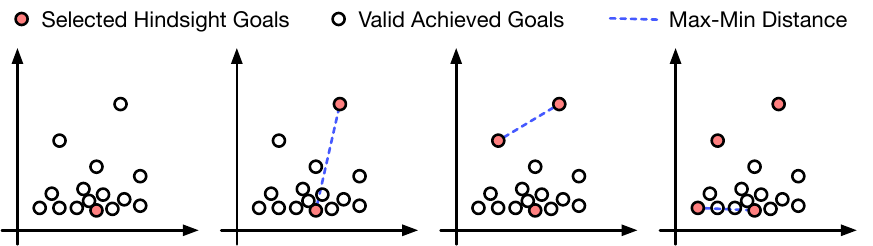}}   
 \vspace{-5pt}     
 \caption{Procedure of Hindsight Goal Filterring} 
 \label{hgf_fig}
 \end{figure}

In hindsight learning, the agent generalizes to reaching the original goal through learning to reach the hindsight goal first.
Therefore, the selection of hindsight goals imposes a direct impact on the performance. 
If the hindsight goals are far from the original ones, the learned policy may not generalize well to the original goals. 
For example, in Fetch PickAndPlace, the initialized random policy barely grasps the target successfully, which results in the hindsight goals majorly distributing on the table. 
Given the original goals up in the air, such a discrepancy can cause a lower learning efficiency. 

In this section, we introduce a heuristic method called Hindsight Goal Filtering(HGF). Intuitively, HGF is trying to filter the most useful goals from the achieved ones instead of random selection. Specifically, based on our analysis (eq. \ref{eq_htrpo_obj}), the performance improves if we reduce the distribution discrepancy between original goals $g$ and hindsight goals $g'$. Ideally, if the distribution of $g'$ matches that of $g$, the agent will reach $g$ after learning to reach $g'$. Therefore, we restrict the selected hindsight goals to distribute in the original goal space whenever possible to cover the area of original goals.

The main idea is shown in Figure \ref{hgf_fig} and the algorithm is summarized in Appendix \ref{ap:HGF}. The input of HGF includes 2 parts: the achieved goal set $\mathcal{G}_a$ and the original goal set $\mathcal{G}_o$. At the beginning, especially for some complex tasks, $\mathcal{G}_a$ can only have small or even no overlap with $\mathcal{G}_o$. Under this situation, we encourage the agent to learn to reach the original goal region by selecting the nearest achieved goals as the hindsight goals. Once some achieved goals fall in the original goal region, they are considered valid achieved goals, and a subset of this intersection will be sampled to cover the region as fully as possible. This subset is selected following the procedure in Figure \ref{hgf_fig}. Note that the distance metric should be determined by the collected original goal distribution. In our experiments, we use the density-weighted Euclidean distance. Specifically, we initialize the hindsight goal set $\mathcal{G}$ with a randomly sampled achieved goal. To make the goal distribute dispersedly, we use Max-Min Distance as the measurement, which indicates the minimal distance between the new goal and the selected ones. By maximizing the minimal distance, it ensures an overall large distance between the new goal and the rest. HGF is related to Curriculum-guided HER (CHER)\cite{fang2019curriculum} to some extent. However, CHER is suitable for transition-based RL, and cannot be applied to episode-based policy gradient algorithms directly.

The complete algorithm of HGF and HTRPO is presented in Appendix \ref{ap:ap_alg}.

\section{Experiments}

Our experiments aims to answer the following questions:
\begin{enumerate}
\item How does HTRPO compared to other methods when performed over diversified tasks?
\item What are the main contributors to HTRPO?
\item How do key parameters affect the performance?
\end{enumerate}

For 1), we show that HTRPO consistently outperforms both HPG and TRPO in the aspect of success rate and sample efficiency in a wide variety of tasks, and achieves state-of-the-art performance in sparse-reward stochastic policy gradient methods. We also provide an in-depth comparison with HER in this part. For 2), we ablate the main components of HTRPO. The ablation study shows that QKL effectively reduces the variance and significantly improves the performance in all tasks. HGF plays a crucial role in improved performance for the more challenging tasks (e.g. Fetch PickAndPlace). For 3), we vary the scale of KL estimation constraint and the numbers of hindsight goals and choose the best parameter settings.

\subsection{Benchmark Settings}

\begin{figure}[t]
\subfigure[Bit Flipping]{\includegraphics[height=2.5cm, width=2.5cm]{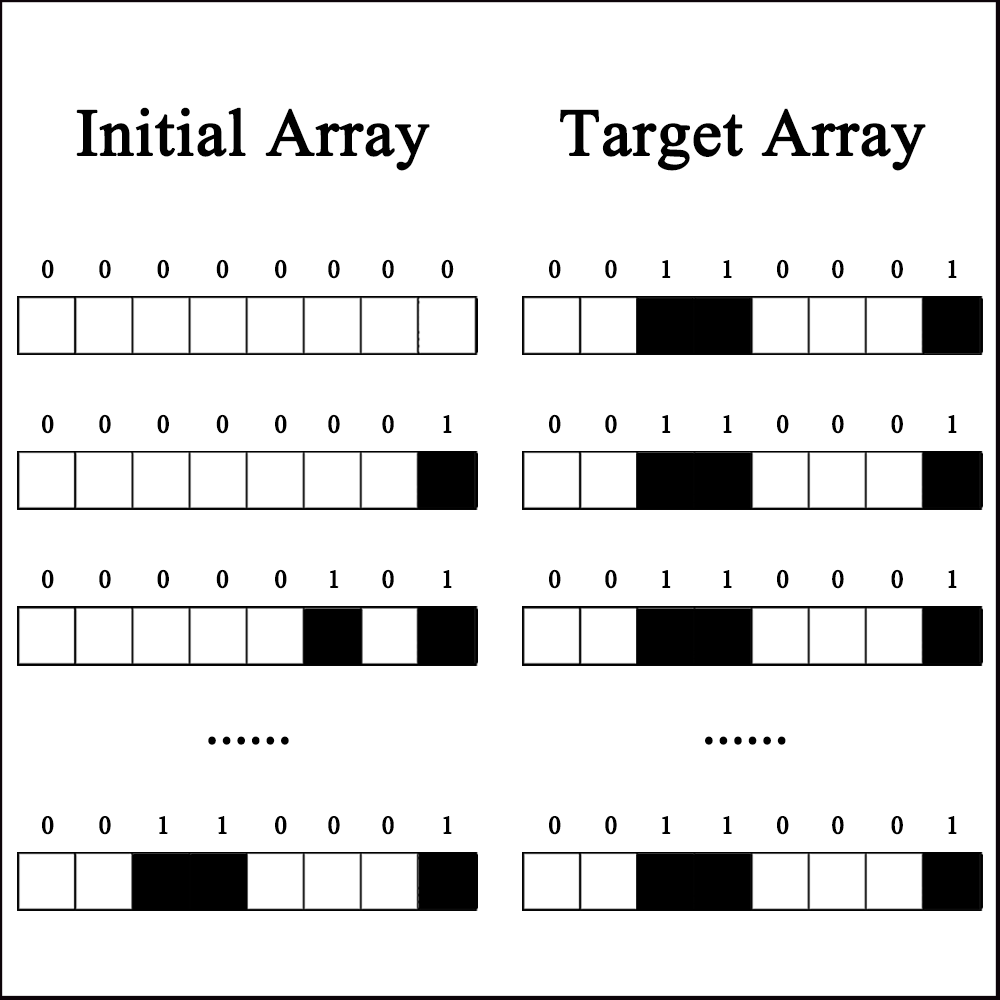}} \
\subfigure[Ms. Pacman]{\includegraphics[height=2.5cm, width=2.5cm]{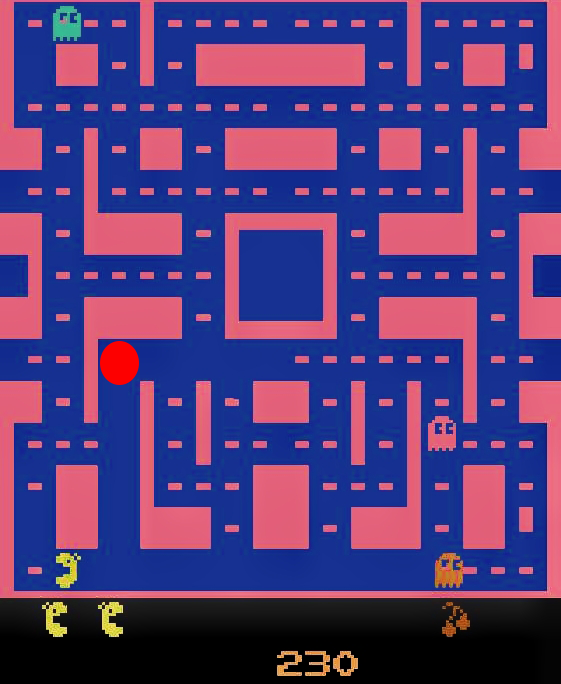}} \
\subfigure[Fetch]{\includegraphics[height=2.5cm, width=2.5cm]{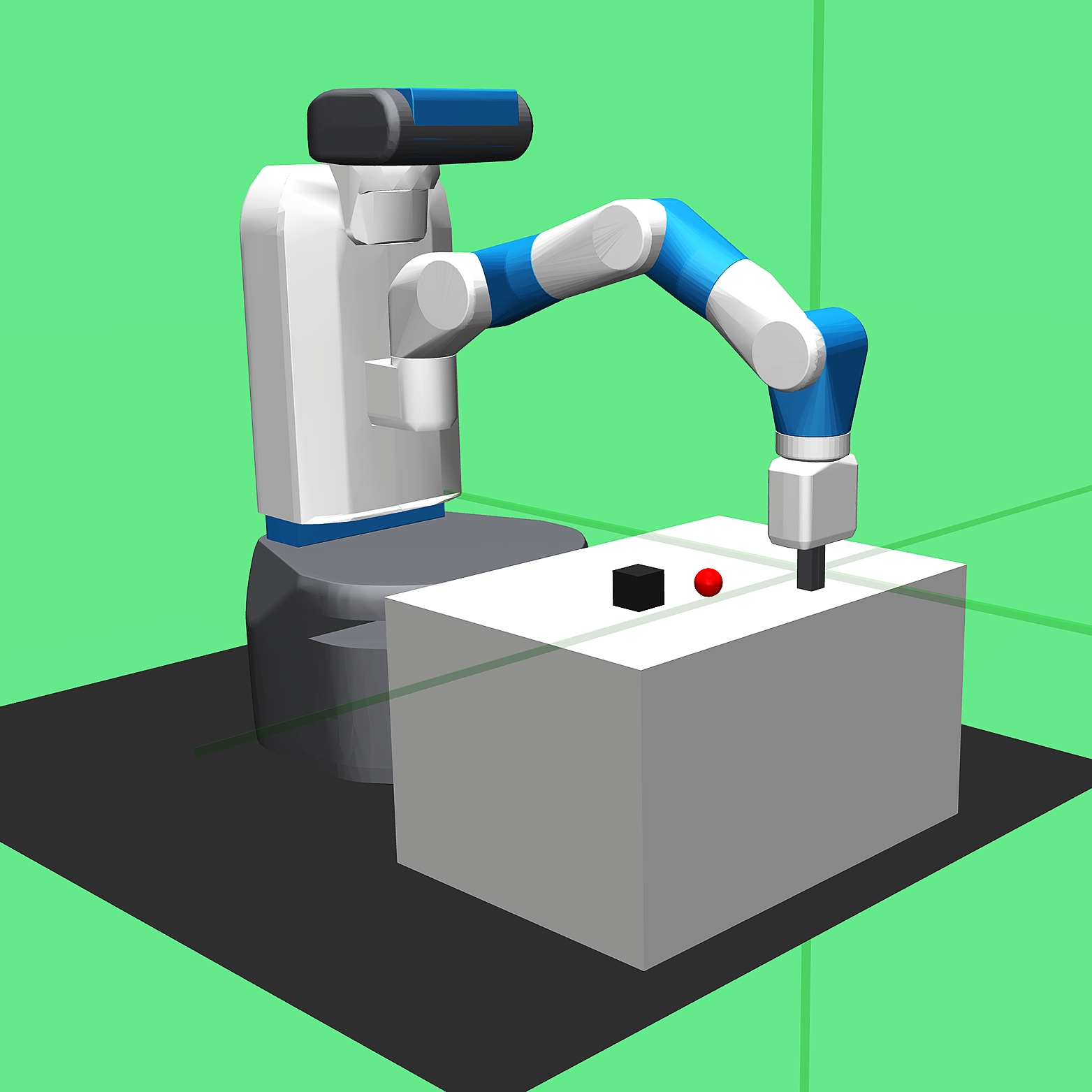}} \
\vspace{-5pt}
\caption{Demonstration of experiment environments}
\label{fig1}
\vspace{-8pt}
\end{figure}

 \begin{figure*}[t]
 \center{\includegraphics[width=0.78\textwidth]{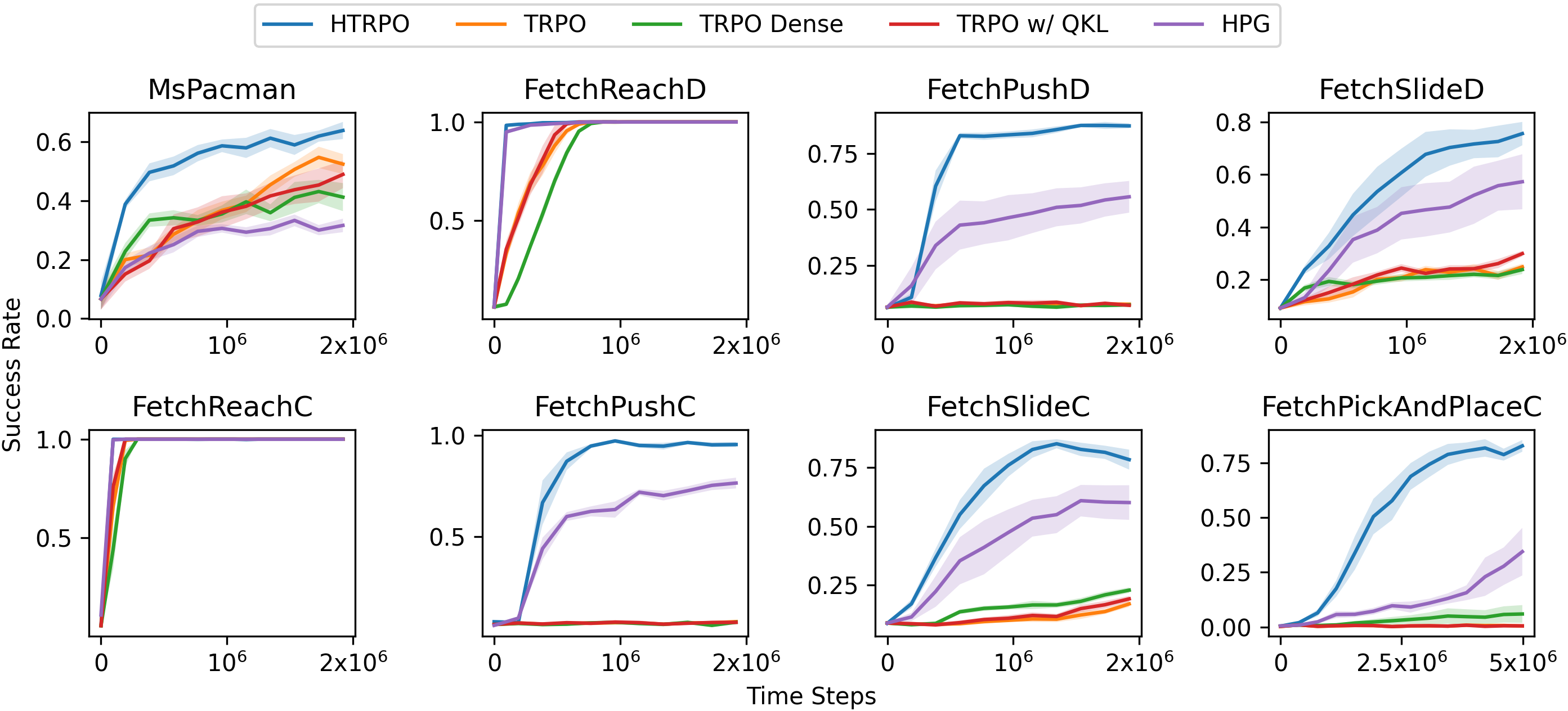}}       
 \caption{Success rate for benchmark environments. {\bf Top row}: performance of discrete environments.  {\bf Bottom row}: performance of continuous environments. The full lines represent the average evaluation over 10 trails and the shaded regions represent the corresponding standard deviation.}  
 \label{robot_res} 
  \vspace{-5pt}
 \end{figure*}

\paragraph{Benchmarks}We implement HTRPO on a variety of sparse reward tasks. Firstly, we test HTRPO in simple benchmarks established in previous work~\cite{andrychowicz2017hindsight} including 4-to-100-Bit Flipping tasks. Secondly, We verify HTRPO's performance in Atari games like Ms. Pac-Man \cite{bellemare13arcade} with complex raw image input to demonstrate its generalization to convolutional neural network policies. Finally, we test HTRPO in simulated robot control tasks like Reach, Push, Slide and PickAndPlace in Fetch~\cite{plappert2018multi} robot environment. As mentioned in \cite{plappert2018multi}, it still remains unexplored that to what extent the policy gradient methods trained with hindsight data can solve continuous control tasks. Since HTRPO is a natural candidate that can be applied to both discrete and continuous tasks, other than discrete Fetch environments introduced in~\cite{rauber2018hindsight}, we also implement HTRPO in continuous environments including Fetch Reach, Fetch Push, Fetch Slide, Fetch PickAndPlace. 

A glimpse of these environments is demonstrated in Figure \ref{fig1}, and the inclusive introductions are included in Appendix \ref{ap:ap9}. Detailed settings of hyperparameters are listed in Appendix \ref{ap:ap10}. All experiments are conducted on a platform with NVIDIA GeForce GTX 1080Ti.

\paragraph{Baselines}We compare HTRPO with HPG \cite{rauber2018hindsight} and TRPO \cite{schulman2015trust}, which are chosen as the baseline algorithms. The reward setting used in our paper is purely sparse reward, i.e., when the task has not been finished, the agent receives 0 reward in each time step, and once the task is finished, the agent will receive a high positive reward. Besides, TRPO is also implemented with dense rewards and the new KL estimation method proposed in Section \ref{cons}. For a fair comparison, we also combine HPG with Hindsight Goal Filtering in our experiments. To demonstrate the performance level of HTRPO more comprehensively, we also compare HTRPO with the well-known HER algorithm. In all experiments, we directly use the accumulated time steps the agent takes while interacting with the environments throughout episodes and batches, and do not count the hindsight steps which are generated using hindsight goals. 

\subsection{Comparative Analysis}


\subsubsection{How does HTRPO compared to other methods when performed over diversified tasks?}

We evaluate HTRPO's performance from success rate and sample efficiency, and test its generality to different tasks including image-based Atari games, and simulated robot control tasks. Results show HTRPO's consistent effectiveness and strong generality to different kinds of tasks and policies.
 
\paragraph{Compare with Baselines}
The success rate curves for the trained policy are demonstrated in Figure \ref{robot_res}. We can conclude that HTRPO consistently outperforms all baselines, including different versions of TRPO and HPG, in most benchmarks, including image-based Atari games (Ms. Pac-Man) and a variety of simulated robot control tasks with different control modes. It demonstrates that HTRPO generalizes well in different kinds of tasks and policies with high-dimensional inputs. Besides, the sample efficiency of HTRPO also exceeds that of HPG, for it reaches a higher average return within less time in most environments. 

\paragraph{Compare with HER} 

We implement HER with DQN~\cite{mnih2015human} for discrete environments and DDPG~\cite{lillicrap2015continuous} for continuous environments based on OpenAI baselines\footnote{https://github.com/openai/baselines}. 
We found that HER cannot work well with the sparse reward setting of HTRPO, i.e., the reward is available only when reaching the goal. Thus, we also follow the reward setting in \cite{andrychowicz2017hindsight} to conduct HER experiments for reference (HER$_{-1}$). 

\textit{Toy Example:} To begin with, we test HTRPO on 4-to-100-Bit Flipping task~\cite{andrychowicz2017hindsight} as well as HER (Figure 4). The maximum training steps are $2\cdot 10^6$. In all Bit Flipping tasks, HTRPO can converge to nearly 100\% success rate with much fewer time steps while HER is much data-inefficient as the number of Bits increases. 

\textit{Benchmarks:} Table \ref{htrpovsher} shows the comparison over the benchmark environments.
We can conclude that: 1) HER can not work quite well with the purely sparse reward setting, and HTRPO outperforms HER in 6 out of 7 benchmarks significantly. 2) For discrete robot control tasks, HTRPO can learn a good policy while HER$_{-1}$+DQN cannot work well. For continuous environments, HER$_{-1}$ slightly outperforms HTRPO.
In summary, HTRPO can be applied both in discrete and continuous tasks without any modification and achieve commendable performance compared to HER.

\subsubsection{What are the main contributors to HTRPO?}

There are mainly 2 components in HTRPO, QKL and HGF, that impose an effect on the performance. Besides, we will also investigate the impact of Weighted Importance Sampling (WIS), which is conducive to variance reduction. To study the effect of reward settings, we implement HTRPO with dense rewards. Selected results are shown in Figure \ref{selectedabla} and the full ablation study is available in Appendix \ref{ap:ap-abla}. We can conclude that: 1) QKL plays a crucial role for the high performance of HTRPO by significantly reducing the estimation variance of KL divergence; 2) HGF can enhance the performance of HTRPO to a higher level; 3) WIS is important since it can reduce the variance of importance sampling significantly; 4) Dense-reward setting harms the performance, which has also been verified in \cite{plappert2018multi}.

\subsubsection{How do key parameters affect the performance?}

\begin{figure}[t]
    \center{\includegraphics[height=3.5cm]{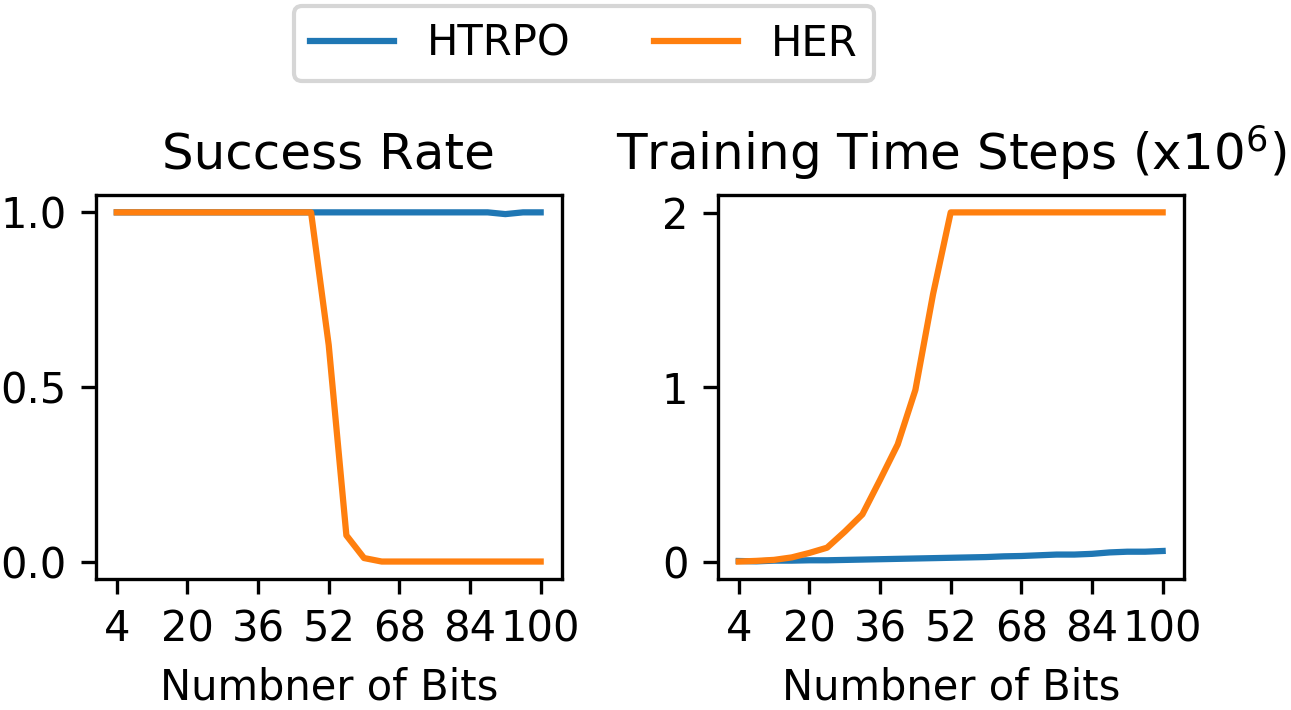}}
    \caption{Performance of Bit Flipping.}
\end{figure}

\begin{table}
\centering
\begin{tabular}{lrrr}
\toprule
Environment  & HER &HER$_{-1}$ & HTRPO \\
\midrule
Ms. Pacman    & {72 $\pm$ 3} &  {\bf 74 $\pm$ 4} & 64 $\pm$ 6 \\
Fetch Reach D  &  53 $\pm$ 41   &{\bf 100 $\pm$ 0}  &  {\bf 100 $\pm$ 0}    \\
Fetch Push D    &   8$\pm$1 &   7$\pm$2   & {\bf 88 $\pm$ 2}\\ 
Fetch Slide D    &  11$\pm$3 &   10$\pm$5  & {\bf 76 $\pm$ 9}\\
Fetch Reach C   & 18$\pm$3 & {\bf 100 $\pm$ 0} &    {\bf 100 $\pm$ 0}  \\
Fetch Push C   & 7$\pm$2 & {\bf 100 $\pm$ 0} & 98 $\pm$ 1  \\
Fetch Slide C   & 1$\pm$1 & \bf 93 $\pm$ 7  & 85 $\pm$ 4   \\
\bottomrule
\end{tabular}
\caption{Success rate comparison between HTRPO and HER (\%). HER$_{-1}$ means using the original -1-and-0 reward setting instead of the purely sparse reward that HTRPO used, i.e., only when the agent achieves the goal can it receive a high reward.}
\label{htrpovsher}
\vspace{-10pt}
\end{table}

We take Continuous Fetch Push as an example to study the impact of different KL estimation constraint scales and different numbers of hindsight goals.

\textbf{Different KL Estimation Constraint Scales:}
KL estimation constraint, i.e. max KL step specifies the trust region, the range within which the agent searches for the next-step optimal policy. In the sense of controlling the scale to which the agent updates the policy per step, this parameter presents similar functionality as learning step size. If set too low, say 5e-6 shown in Figure 6, it would inevitably slow down the converging speed. If set too high, the potentially large divergence between the new and old policy may violate the premise for some core parts of HTRPO theory derivation including Theorem 3.3 and HTRPO solving process.

\begin{figure}[t]
\center{\includegraphics[width=0.45\textwidth]{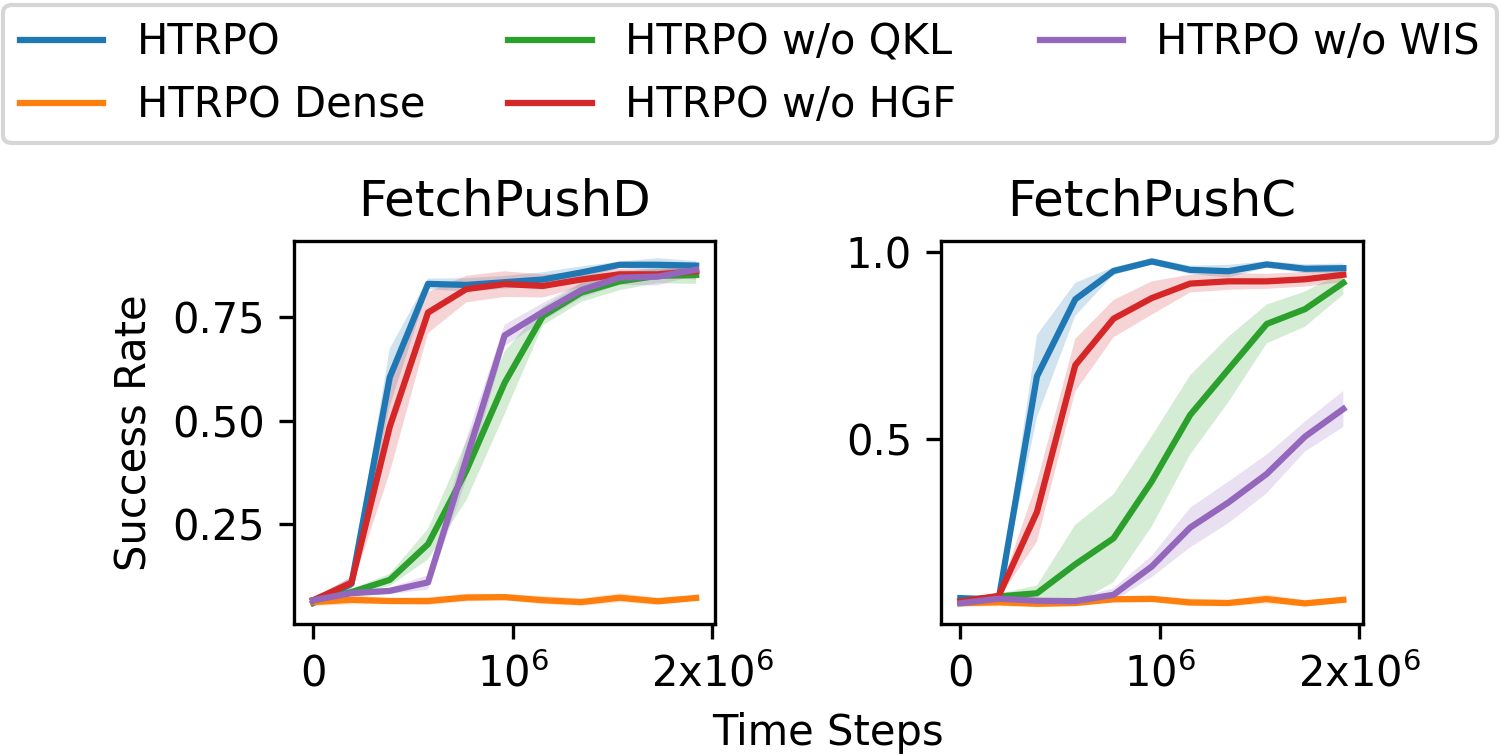}}
\caption{Ablation Experiments.}
\label{selectedabla} 
\end{figure}

\begin{figure}[t]
\vspace{-7.5pt}
\begin{tabular}{cc}
\begin{minipage}[t]{0.5\linewidth}\label{hyperparamskl}
    \includegraphics[height=3.8cm]{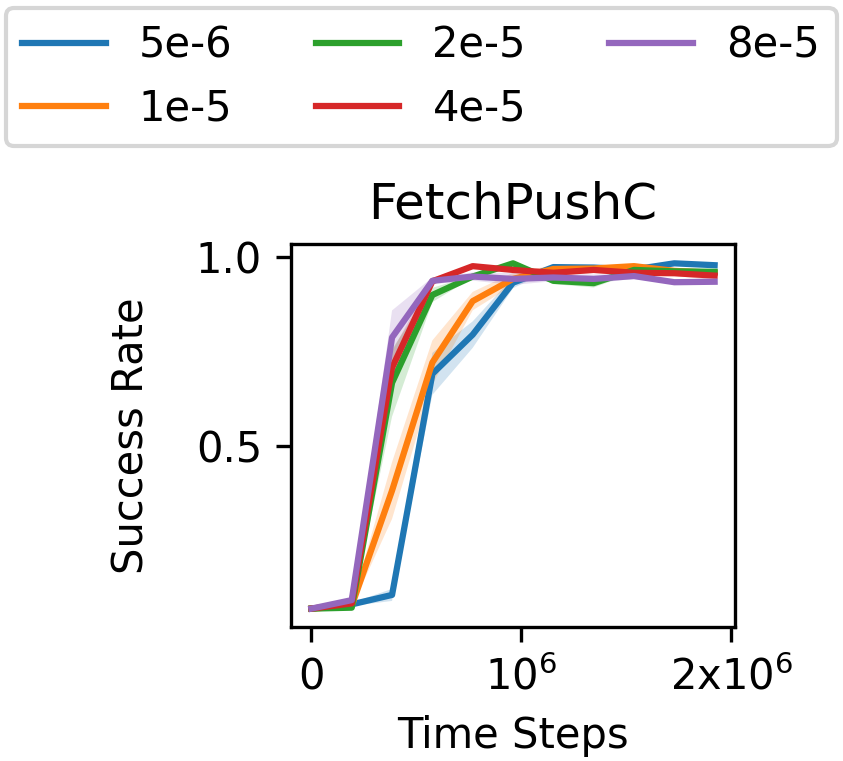}
    \caption{Max KL steps.}
\end{minipage}
\begin{minipage}[t]{0.5\linewidth}\label{hyperparamsgnum}
    \includegraphics[height=3.8cm]{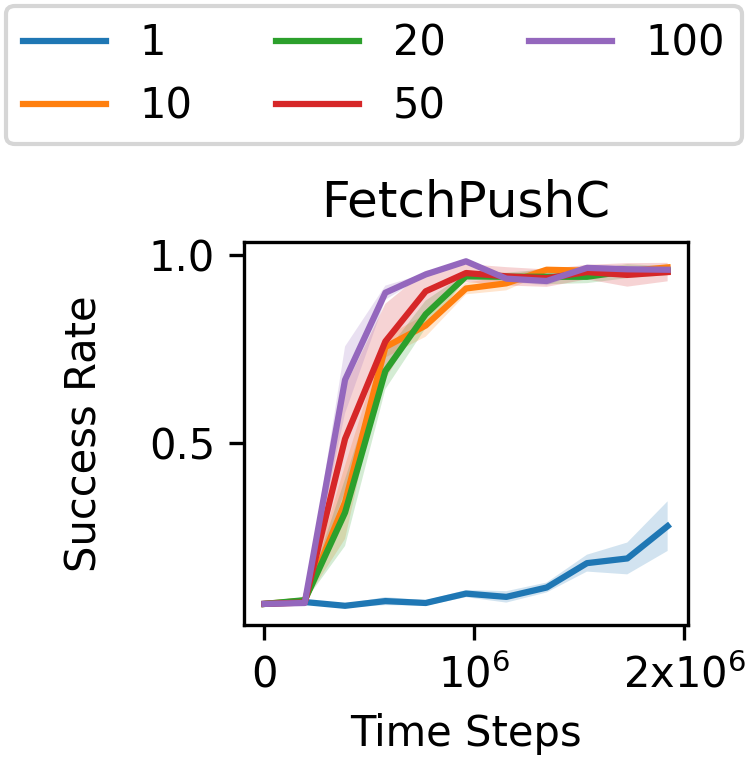}
    \caption{Goal numbers.}
\end{minipage}
\end{tabular}
\vspace{-10pt}
\end{figure}

\textbf{Different Number of Hindsight Goals:}
From the results in Figure 7, it is straightforward that more hindsight goals lead to faster converging speed. This phenomenon accords with the mechanism of how hindsight methodology deals with sparse reward scenarios, i.e. it augments the sample pool with substantial hindsight data rather than leaving it with few valid original trajectories. It's intuitive that the more hindsight data there are, the higher sample efficiency HTRPO achieves. However, limited by the hardware resources, we need to trade off the sampled goal number.

\section{Conclusion}

We proposed Hindsight Trust Region Policy Optimization(HTRPO), a new RL algorithm that extends the highly successful TRPO algorithm with hindsight to tackle the challenge of sparse rewards. We show that with the help of the proposed Quadratic KL divergence Estimation (QKL), HTRPO significantly reduces the variance of KL estimation and improves the performance and learning stability. Moreover, we design a Hindsight Goal Filtering mechanism to narrow the discrepancy between hindsight and original goal space, leading to better performance. Results on diversified benchmarks demonstrate the effectiveness of HTRPO.

Since HTRPO is a natural candidate for both discrete and continuous tasks and the QKL constraint gets rid of the demand for analytical form, it is promising to optimize policies with non-Gaussian (e.g. GMM) or mixed (discrete$+$continuous) action space. It also provides the possibility to tackle high-dimensional real-world problems and train robot control policies without arduous reward shaping. Besides, HGF can be integrated into hindsight-goal exploration methods naturally~\cite{ren2019exploration,pitis2020maximum}, which should lead to a higher performance.

\section*{Acknowledgements}

This work was supported in part by NSFC under grant No.91748208, No.62088102, No.61973246, Shaanxi Project under grant No.2018ZDCXLGY0607, and the program of the Ministry of Education. D. Hsu is supported by the National Research Foundation, Singapore under its AI Singapore Program (AISG Award No: AISG2-RP-2020-016).

\bibliographystyle{named}
\bibliography{ijcai21}

\newpage

\appendix

\section{Proof of Quadratic KL Estimation Method}

\subsection{Proposition 1}\label{ap:ap5}
\begin{lamma} \label{ap:lamma1}
Given two distibutions $p(x)$ and $q(x)$, $q(x) = p(x) + \eta(x)$, in which $\eta(x)$ is the variation of $q(x)$ at $p(x)$.
\begin{align}
&\mathop{\mathbb{E}} \left[ \log p(x) - \log q(x) \right ] \nonumber \\
&= \mathop{\mathbb{E}} \left[ \frac{1}{2} ( \log p(x) - \log q(x))^2 \right] + \mathop{\mathbb{E}} \left[ o(\eta(x)^3 )  \right] \label{ap:eqc.1}
\end{align}
\end{lamma}

\proof Consider the second order Taylor expansion of $\log q(x)$ at $p(x)$,
\begin{align}
\log q(x) = \log p(x) + \frac{1}{p(x)} \eta(x) - \frac{1}{2p(x)^2} \eta(x)^2 + o(\eta(x)^3)
\end{align}
For the left side of eq.\ref{ap:eqc.1},
\begin{align} 
&\mathop{\mathbb{E}} \left[ \log p(x) - \log q(x) \right ] \nonumber \\
&= \mathop{\mathbb{E}} \left[ - \frac{1}{p(x)} \eta(x) + \frac{1}{2p(x)^2} \eta(x)^2 - o(\eta(x)^3) \right ]  \nonumber \\
&= \int (- \frac{1}{p(x)} \eta(x) + \frac{1}{2p(x)^2} \eta(x)^2 - o(\eta(x)^3)) p(x) \, dx  \nonumber \\
&= \int (\frac{1}{2p(x)} \eta(x)^2 - p(x) o(\eta(x)^3) ) \, dx.
\end{align}
For the first term on the right side of eq.\ref{ap:eqc.1},
\begin{align} \nonumber
&\mathop{\mathbb{E}} \left[ \frac{1}{2} ( \log p(x) - \log q(x))^2 \right] \\ \nonumber 
&= \frac{1}{2} \mathop{\mathbb{E}} \left[ (- \frac{1}{p(x)} \eta(x) + \frac{1}{2p(x)^2} \eta(x)^2 - o(\eta(x)^3))^2 \right ] \\ \nonumber
&= \frac{1}{2} \mathop{\mathbb{E}} \left[ \frac{1}{p(x)^2} \eta(x)^2 + o(\eta(x)^3) \right ] \\ \nonumber
&= \frac{1}{2} \int (\frac{1}{p(x)^2} \eta(x)^2 + o(\eta(x)^3)) p(x) \, dx \\ \nonumber
&= \int (\frac{1}{2p(x)} \eta(x)^2 + \frac{1}{2} p(x) o(\eta(x)^3)) \, dx \\ \nonumber
&= \int (\frac{1}{2p(x)} \eta(x)^2 - p(x) o(\eta(x)^3) ) \, dx + \int p(x) o(\eta(x)^3) \, dx \\
&= \mathop{\mathbb{E}} \left[ \log p(x) - \log q(x) \right ] + \mathop{\mathbb{E}} \left[ o(\eta(x)^3 )  \right].
\end{align}

\paragraph{Proposition 1.} (Approximation of Constraint Function). For policy $\pi_{\tilde{\theta}}(a|s)$ and $\pi_{\theta}(a|s)$, and for $\eta = \pi_{\theta}(a|s) - \pi_{\tilde{\theta}}(a|s)$,
\begin{align} \label{ap:eq3.2}
&\mathop{\mathbb{E}}_{a \sim \pi_{\tilde{\theta}}(a|s)} \left[ \log \pi_{\tilde{\theta}}(a|s) - \log \pi_{\theta}(a|s) \right ] \nonumber \\
= & \ \mathop{\mathbb{E}}_{a \sim \pi_{\tilde{\theta}}(a|s)} \left[ \frac{1}{2} ( \log \pi_{\tilde{\theta}}(a|s) - \log \pi_{\theta}(a|s))^2 \right] \nonumber \\
&+ \mathop{\mathbb{E}}_{a \sim \pi_{\tilde{\theta}}(a|s)} \left[ o( \eta^3 )  \right].
\end{align}

\proof Based on Lemma \ref{ap:lamma1}, let $p(x) = \pi_{\tilde{\theta}}(a|s)$ and $q(x)=\pi_{\theta}(a|s)$, eq.\ref{ap:eqc.1} results in eq.\ref{ap:eq3.2}.

\
\subsection{Proposition 2} \label{ap:ap6}
\begin{lamma} \label{ap:lamma2}
For any random variable $Y \in \left[ 0, 0.5 \right ]$,
\begin{align}
\mathrm{Var} ( Y^2 ) \leq \mathrm{Var} ( Y ),
\end{align}
in which $\mathrm{Var}(Y)$ denotes the variance of $Y$.
\end{lamma}

\proof
\begin{align} \nonumber
&\mathrm{Var} ( Y ) - \mathrm{Var} ( Y^2 ) \\ \nonumber
&= \mathop{\mathbb{E}}(Y^2) - \left[ \mathop{\mathbb{E}}(Y) \right]^2 - \mathop{\mathbb{E}}(Y^4) +  \left[ \mathop{\mathbb{E}}(Y^2) \right]^2 \\ \nonumber
&= \left[ \mathop{\mathbb{E}} (Y^2) - \mathop{\mathbb{E}}(Y^4) \right] - \left[ \left[ \mathop{\mathbb{E}}(Y) \right]^2 - \left[ \mathop{\mathbb{E}}(Y^2) \right]^2 \right] \\ \nonumber
&= \mathop{\mathbb{E}} \left[ Y^2(1-Y)(1+Y) \right] - \mathop{\mathbb{E}} \left[ Y(1-Y)\right] \mathop{\mathbb{E}} \left[ Y(1+Y)\right] \\
&= \mathrm{Cov}(Y(1+Y), Y(1-Y)).
\end{align}
Denote $X_1(Y) = Y(1+Y)$ and $X_2(Y) = Y(1-Y)$. Then,
\begin{align} \nonumber
\mathrm{Var} ( Y ) - \mathrm{Var} ( Y^2 ) &= \mathrm{Cov}(X_1, X_2) \\
&= \mathop{\mathbb{E}} \left[ X_1 \left(	X_2 - \mathop{\mathbb{E}} (X_2) \right) \right]
\end{align}
There always exists $Y_0 \in \left[ 0, 0.5 \right] $ that satisfies $X_2(Y_0) = \mathop{\mathbb{E}}(X_2)$. When $Y = Y_0$, let $X_1(Y_0) = \mu_1$ in which $\mu_1 $ is a constant. Then the equation can be converted by the following steps:
\begin{align}\nonumber
&\mathrm{Var} ( Y ) - \mathrm{Var} ( Y^2 ) \\ \nonumber
&= \mathop{\mathbb{E}} \left[ \left( X_1 - \mu_1 \right) \left(	X_2 - \mathop{\mathbb{E}} (X_2) \right) \right] + \mu_1 \mathop{\mathbb{E}} \left[ X_2 - \mathop{\mathbb{E}} (X_2) \right] \\
&= \mathop{\mathbb{E}} \left[ \left( X_1 - \mu_1 \right) \left(	X_2 - \mathop{\mathbb{E}} (X_2) \right) \right] \label{ap:eqc.2}
\end{align}

Thus, when $Y = Y_0$, the two factors in eq.\ref{ap:eqc.2}, $\left( X_1 - \mu_1 \right)$ and $ \left( X_2 - \mathop{\mathbb{E}} (X_2) \right)$ equal to 0 simultaneously.

Also, it is easy to notice that when $Y \in \left[ 0, 0.5 \right]$, $X_1$ and $X_2$ are strictly increasing with the increase of $Y$. Thus, $\left( X_1 - \mu_1 \right)$ and $ \left( X_2 - \mathop{\mathbb{E}} (X_2) \right)$ are either both positive or both negative, if not zero. Therefore,
\begin{align}
\mathrm{Var} ( Y ) - \mathrm{Var} ( Y^2 ) \geq 0.
\end{align}

\begin{lamma} \label{ap:lamma3}
For any random variable Y,
\begin{align}
\mathrm{Var} ( | Y | ) \leq \mathrm{Var} ( Y ), \label{ap:eqc.3}
\end{align}
in which $\mathrm{Var}(Y)$ denotes the variance of $Y$.
\end{lamma}

\proof Apparently,
\begin{align}
\int_y | f(y) | \,dy \geq |\int_y f(y) \,dy|.
\end{align}
Consequently,
\begin{align}
\mathop{\mathbb{E}}(|Y|) \geq | \mathop{\mathbb{E}}(Y) |.
\end{align}
For that
\begin{align}
\mathop{\mathrm{Var}}(Y) = \mathop{\mathbb{E}}(Y^2) - (\mathop{\mathbb{E}}(Y))^2,
\end{align}
we have
\begin{align}
\mathrm{Var} ( | Y | ) \leq \mathrm{Var} ( Y ).
\end{align}

\paragraph{Proposition 2.} (Variance of Constraint Function). For policy $\pi_{\tilde{\theta}}(a|s)$ and $\pi_{\theta}(a|s)$, let $\mathrm{Var}$ denotes the variance of a variable. When $\log \pi_{\tilde{\theta}}(a|s) - \log \pi_{\theta}(a|s) \in \left[ -0.5 , 0.5 \right ]$, then
\begin{align}
\mathop{\mathrm{Var}}_{a \sim \pi_{\tilde{\theta}}(a|s)} \left[ \frac{( \log \pi_{\tilde{\theta}}(a|s) - \log \pi_{\theta}(a|s))^2}{2}  \right] \nonumber \\
< \mathop{\mathrm{Var}}_{a \sim \pi_{\tilde{\theta}}(a|s)} \left[ \log \pi_{\tilde{\theta}}(a|s) - \log \pi_{\theta}(a|s) \right].
\end{align}

\proof \ let $Y = | \log \pi_{\tilde{\theta}}(a|s) - \log \pi_{\theta}(a|s) | $. Given Lemma \ref{ap:lamma2},
\begin{align}
&\mathop{\mathrm{Var}}_{a \sim \pi_{\tilde{\theta}}(a|s)} \left[ | \log \pi_{\tilde{\theta}}(a|s) - \log \pi_{\theta}(a|s)|^2 \right] \nonumber \\
&\leq \mathop{\mathrm{Var}}_{a \sim \pi_{\tilde{\theta}}(a|s)} \left[ | \log \pi_{\tilde{\theta}}(a|s) - \log \pi_{\theta}(a|s) | \right]  \label{ap:eqc.4}
\end{align}
Given Lemma \ref{ap:lamma3},
\begin{align}
&\mathop{\mathrm{Var}}_{a \sim \pi_{\tilde{\theta}}(a|s)} \left[ | \log \pi_{\tilde{\theta}}(a|s) - \log \pi_{\theta}(a|s) | \right] \nonumber \\
&\leq \mathop{\mathrm{Var}}_{a \sim \pi_{\tilde{\theta}}(a|s)} \left[ \log \pi_{\tilde{\theta}}(a|s) - \log \pi_{\theta}(a|s) \right] . \label{ap:eqc.5}
\end{align}
With the transitivity of inequality, combining eq.\ref{ap:eqc.4} and eq.\ref{ap:eqc.5}, we know that
\begin{align}
&\mathop{\mathrm{Var}}_{a \sim \pi_{\tilde{\theta}}(a|s)} \left[ | \log \pi_{\tilde{\theta}}(a|s) - \log \pi_{\theta}(a|s) |^2 \right] \nonumber \\
&\leq \mathop{\mathrm{Var}}_{a \sim \pi_{\tilde{\theta}}(a|s)} \left[ \log \pi_{\tilde{\theta}}(a|s) - \log \pi_{\theta}(a|s) \right].
\end{align}

\section{Derivation of QKL-TRPO} \label{ap:ap-ca}

\subsection{Proof of Proposition 3}

\begin{lamma} \label{ap:lamma4}
 (Policy Improvement Guarantee of TRPO) Given two policies $\pi_{\theta}$ and $\pi_{\tilde{\theta}}$, we have:
\begin{align}
\eta(\pi_{\theta}) \geq  L_{\pi_{\tilde{\theta}}}(\pi_{\theta}) - C\alpha^2
\end{align}
where 
\begin{align}
L_{\pi_{\tilde{\theta}}}(\pi_{\theta}) = \eta(\pi_{\tilde{\theta}}) + E_{s\sim \pi_{\tilde{\theta}}(s), a\sim \pi_{\theta}(a|s)}[A_{\pi_{\tilde{\theta}}}(s, a)]
\end{align}
and $\eta(\pi_{\theta}) =E\left[\sum \gamma^t r_t\right]$, $C=\frac{4\beta\gamma}{(1-\gamma)^2}$, $\beta=max_{s,a}|A_{\pi_{\tilde{\theta}}} (s,a)|$, $\alpha = D_{TV}^{max}(\pi_{\tilde{\theta}}(a|s), \pi_{\theta}(a|s))$. 
\end{lamma}

\proof
The detailed proof can be found in TRPO\cite{schulman2015trust}.

\paragraph{Proposition 3.}\label{ap:propos3}
(Policy Improvement Guarantee) Given two policies $\pi_{\theta}$ and $\pi_{\tilde{\theta}}$, Let 
\begin{align}
s^*=\arg\max_s D_{TV}(\pi_{\tilde{\theta}}(a|s), \pi_{\theta}(a|s)) \nonumber
\end{align}
If $\frac{D_{KL}(\pi_{\tilde{\theta}}(a|s^*), \pi_{\theta}(a|s^*))} {D_{QKL}(\pi_{\tilde{\theta}}(a|s^*), \pi_{\theta}(a|s^*))} \leq \frac{2}{\ln 2}$, then 
\begin{align}
\eta(\pi_{\theta}) \geq  L_{\pi_{\tilde{\theta}}}(\pi_{\theta}) - CD_{QKL}^{max}(\pi_{\tilde{\theta}}(a|s), \pi_{\theta}(a|s))
\end{align}
where 
\begin{align}
L_{\pi_{\tilde{\theta}}}(\pi_{\theta}) = \eta(\pi_{\tilde{\theta}}) + E_{s\sim \pi_{\tilde{\theta}}(s), a\sim \pi_{\theta}(a|s)}[A_{\pi_{\tilde{\theta}}}(s, a)]
\end{align}
and $\eta(\pi_{\theta}) =E\left[\sum \gamma^t r_t\right]$ is the expected return, $C=\frac{4\beta\gamma}{(1-\gamma)^2}$, $\beta=max_{s,a}|A_{\pi_{\tilde{\theta}}} (s,a)|$, $D_{TV}(p,q) = \frac{1}{2}\sum_i\left|p_i-q_i\right|$.

\proof From Pinsker's inequality, we have
\begin{align}
D_{TV}^2(\pi_{\tilde{\theta}}, \pi_{\theta}) = \frac{1}{4}\left\| \pi_{\tilde{\theta}} - \pi_{\theta} \right\|_1^2 \leq \frac{\ln 2}{2} D_{KL}(\pi_{\tilde{\theta}}, \pi_{\theta})
\end{align}
Given $\frac{D_{KL}(\pi_{\tilde{\theta}}(a|s^*), \pi_{\theta}(a|s^*))} {D_{QKL}(\pi_{\tilde{\theta}}(a|s^*), \pi_{\theta}(a|s^*))} \leq \frac{2}{\ln 2}$, we have 
\begin{align}
D_{TV}^2(\pi_{\tilde{\theta}}(a|s^*), \pi_{\theta}(a|s^*))  \leq D_{QKL}(\pi_{\tilde{\theta}}(a|s^*), \pi_{\theta}(a|s^*))
\end{align}
Given Lemma \ref{ap:lamma4}, we have proved Proposition \ref{ap:propos3}.

\subsection{Derivation of the Practical Algorithm of QKL-TRPO}\label{ap:ap-pracqkltrpo}


To guarantee policy improvement, according to Proposition \ref{ap:propos3}, we can optimize the policy by iteratively solving the following problem (also shown in Algorithm \ref{ap:alg-MonoConvAlg}:
\begin{equation} \pi^*=\arg\max_{\pi_\theta}\left[L_{\pi_{\tilde{\theta}}}(\pi_\theta) -CD_{QKL}^{max}(\pi_{\tilde{\theta}}(a|s), \pi_\theta(a|s)) \right] \end{equation}
\begin{equation} s.t. \quad s^* = \arg\max_{s}D_{TV}(\pi_{\tilde{\theta}}(a|s), \pi_\theta(a|s)) \end{equation}
\begin{equation} \quad \frac{D_{KL}(\pi_{\tilde{\theta}}(a|s^*), \pi_\theta(a|s^*))}{D_{QKL}(\pi_{\tilde{\theta}}(a|s^*), \pi_\theta(a|s^*))} \leq \frac{2}{\ln 2}\end{equation}
where $ L_{\pi_{\tilde{\theta}}}(\pi_\theta) = \sum_{s}\rho_{\pi_{\tilde{\theta}}}\sum_{a}\pi_\theta(a|s)A_{\pi_{\tilde{\theta}}}(s,a)$. Such an optimization results in a monotonically improvement of the policy in each optimization step since:
\begin{align}
\eta(\pi^*) &\geq L_{\pi_{\tilde{\theta}}}(\pi^*) -CD_{QKL}^{max}(\pi_{\tilde{\theta}}(a|s), \pi^*(a|s)) \nonumber \\
& \geq L_{\pi_{\tilde{\theta}}}(\pi_{\tilde{\theta}}) -CD_{QKL}^{max}(\pi_{\tilde{\theta}}(a|s), \pi_{\tilde{\theta}}(a|s)) \nonumber \\
& = L_{\pi_{\tilde{\theta}}}(\pi_{\tilde{\theta}}) = \eta(\pi_{\tilde{\theta}})
\end{align}
As suggested in TRPO, to ensure a reasonable learning step size, we transform the penalty on $D_{QKL}^{max}$ to a trust region constraint:
\begin{equation} \pi^*=\arg\max_{\pi_\theta}L_{\pi_{\tilde{\theta}}}(\pi_\theta) \end{equation}
\begin{equation} \label{ap:eq-tvs} s.t. \quad s^* = \arg\max_{s}D_{TV}(\pi_{\tilde{\theta}}(a|s), \pi_\theta(a|s)) \end{equation}
\begin{equation} \label{ap:eq-conspd} \quad \frac{D_{KL}(\pi_{\tilde{\theta}}(a|s^*), \pi_\theta(a|s^*))}{D_{QKL}(\pi_{\tilde{\theta}}(a|s^*), \pi_\theta(a|s^*))} \leq \frac{2}{\ln 2}\end{equation}
\begin{equation} \quad \label{ap:eq-consmaxqkl} D_{QKL}^{max}(\pi_{\tilde{\theta}}(a|s), \pi_\theta(a|s)) \leq \epsilon_1 \end{equation}
However, to solve such a constrained optimization problem, there are three difficulties: 1) the constraint represented by eq. \ref{ap:eq-conspd} is hard to deal with; 2) it is difficult to find the point $s^*$ that maximizes the TV divergence; 3) Eq. \ref{ap:eq-consmaxqkl} imposes a constraint on $D_{QKL}$ at every state, which is intractable due to the large state space.

\begin{algorithm}[h]
\caption{A Monotonically-Converging Algorithm for Policy Optimization}
\label{ap:alg-MonoConvAlg}
\begin{algorithmic}
\STATE Initialize policy $\pi_0$;
\STATE $i=0$;
\WHILE{$\pi_i$ does not converge}
\STATE Compute all $A_{\pi_i}(s,a)$;
\STATE Solve the constrained optimization problem
\STATE 
\begin{equation} \nonumber
\pi^*=\arg\max_{\pi}\left[L_{\pi_{i}}(\pi) -CD_{QKL}^{max}(\pi_{i}(a|s), \pi(a|s)) \right]
\end{equation}
\STATE \begin{equation} \nonumber s.t. \quad s^* = \arg\max_{s}D_{TV}(\pi_{i}(a|s), \pi(a|s)) \end{equation}
\STATE \begin{equation} \nonumber \quad \frac{D_{KL}(\pi_{i}(a|s^*), \pi(a|s^*))}{D_{QKL}(\pi_{i}(a|s^*), \pi(a|s^*))} \leq \frac{2}{\ln 2}\end{equation}
\STATE \begin{equation} \nonumber \quad L_{\pi_i}(\pi) = \eta(\pi_i) + E_{s\sim \pi_i(s), a\sim \pi(a|s)}[A_{\pi_i}(s, a)] \end{equation}
\STATE $i=i+1$;
\STATE $\pi_i = \pi^*$;
\ENDWHILE
\STATE {\bf Return} $\pi^*$
\end{algorithmic}
\end{algorithm}

To tackle the first two challenges, we constrain the $D_{QKL}$ instead of the eq. \ref{ap:eq-conspd}, that is
\begin{equation} \label{ap:eq-consqkl2}D_{QKL}(\pi_{\tilde{\theta}}(a|s^*), \pi_\theta(a|s^*))\leq \epsilon_2\end{equation} 
Such a constraint is based on that $\lim_{D_{KL} \to 0} \frac{D_{KL}}{D_{QKL}}=\lim_{D_{QKL} \to 0} \frac{D_{KL}}{D_{QKL}}=1$, i.e., for $\forall$ positive $\delta$, there exists a positive threshold $\epsilon_2$, when $D_{QKL}\leq \epsilon_2$, $\left|\frac{D_{KL}}{D_{QKL}} - 1\right|\leq \delta$. When $\delta = \frac{2}{\ln 2} - 1$, we have:
\begin{equation} \label{ap:eq-conspd2} 2-\frac{2}{\ln 2} \leq \frac{D_{KL}(\pi_{i}(a|s^*), \pi(a|s^*))}{D_{QKL}(\pi_{i}(a|s^*), \pi(a|s^*))} \leq \frac{2}{\ln 2}\end{equation} 
which results in a stronger constraint compared with eq. \ref{ap:eq-conspd}. Intuitively, such a constraint intends to make the two policies not far from each other at a certain point $s^*$. Since it is hard to obtain the precise value of $\epsilon_2$, we set it to be a hyperparameter which controls the distance between $\pi_{\tilde{\theta}}$ and $\pi_{\theta}$. After that, because 
\begin{equation} D_{QKL}(\pi_{\tilde{\theta}}(a|s^*), \pi_\theta(a|s^*)) \leq D_{QKL}^{max}(\pi_{\tilde{\theta}}(a|s), \pi_\theta(a|s)) \end{equation}
obviously holds, we can further introduce a tighter constraint than eq. \ref{ap:eq-consqkl2}:
\begin{equation} \label{ap:eq-consmaxqkl2}D_{QKL}^{max}(\pi_{\tilde{\theta}}(a|s), \pi_\theta(a|s))\leq \epsilon_2\end{equation} 
which is basically the same with eq. \ref{ap:eq-consmaxqkl} except for the constant threshold. By taking $\epsilon=\min(\epsilon_1,\epsilon_2)$, we can merge eq. \ref{ap:eq-consmaxqkl} and \ref{ap:eq-consmaxqkl2}:
\begin{equation} \pi^*=\arg\max_{\pi_\theta}L_{\pi_{\tilde{\theta}}}(\pi_\theta) \end{equation}
\begin{equation} \label{ap:eq-finalmaxqlcons}s.t. \quad  D_{QKL}^{max}(\pi_{\tilde{\theta}}(a|s), \pi_\theta(a|s)) \leq \epsilon \end{equation}

To tackle the third challenge, inspired by TRPO\cite{schulman2015trust}, we use a heuristically approximate constraint $E_s\left[D_{QKL}(\pi_{\tilde{\theta}}(a|s), \pi_\theta(a|s))\right] \leq \epsilon$ instead of eq. \ref{ap:eq-finalmaxqlcons}:
\begin{equation} \pi^*=\arg\max_{\pi_\theta}L_{\pi_{\tilde{\theta}}}(\pi_\theta) \end{equation}
\begin{equation} s.t. \quad  \mathop{\mathbb{E}}_s\left[D_{QKL}(\pi_{\tilde{\theta}}(a|s), \pi_\theta(a|s))\right] \leq \epsilon \end{equation}
This problem formulation is equivalent to:
\begin{align} \nonumber
\max_{\theta} \ \mathop{\mathbb{E}}_{s, a \sim \rho_{\tilde{\theta}}(s,a)} \left [ \frac{\pi_{\theta}(a|s)}{\pi_{\tilde{\theta}}(a|s)} A_{\tilde{\theta}}(s, a) \right ]
\end{align}
\vspace{-10pt}
\begin{align}\nonumber
s.t. \ \mathop{\mathbb{E}}_{s, a} \left[\frac{1}{2}( \log \pi_{\tilde{\theta}}(a|s) - \log \pi_{\theta}(a|s))^2 \right] \leq \epsilon
\end{align}
which is called QKL-TRPO.

\section{Proof for Hindsight Trust Region Optimization}

\subsection{Reformulation of TRPO Objective}  \label{ap:ap1}

\begin{align}
&\max_{\theta} \ \mathop{\mathbb{E}}_{s, a \sim \rho_{\tilde{\theta}}(s,a)} \left [ \frac{\pi_{\theta}(a|s)}{\pi_{\tilde{\theta}}(a|s)} A_{\tilde{\theta}}(s, a) \right ] \label{ap:eq1.2} \\
&=\max_{\theta} \ \mathop{\mathbb{E}}_{\tau \sim p_{\tilde{\theta}}(\tau)} \left [ \sum_{t=0}^{\infty} \gamma^t \frac{\pi_{\theta}(a_{t}|s_{t})}{\pi_{\tilde{\theta}}(a_{t}|s_{t})} A_{\tilde{\theta} }(s_t, a_t) \right ] 
\end{align}

\proof
With no influence to the optimal solution, we can multiply eq.\ref{ap:eq1.2} by a constant $\frac{1}{1-\gamma}$,
\begin{align}\nonumber
&L_{\tilde{\theta}}(\theta)= \frac{1}{1-\gamma} \mathop{\mathbb{E}}_{s \sim \rho_{\tilde{\theta}},a \sim \pi_{\tilde{\theta}}(a \mid s)} \left [ \frac{\pi_{\theta}(a|s)}{\pi_{\tilde{\theta}}(a|s)} A_{\tilde{\theta} }(s, a) \right ] \\ \nonumber
&= \frac{1}{1-\gamma}  \sum_{s \in \mathcal{S}}  \frac{\sum_{t=0}^{\infty}\gamma^t P(s_t=s)}{\frac{1}{1-\gamma}}  \mathop{\mathbb{E}}_{ a \sim \pi_{\tilde{\theta}}(a|s)} \left [  \frac{\pi_{\theta}(a|s)}{\pi_{\tilde{\theta}}(a|s)} A_{\tilde{\theta} }(s, a) \right ] \\ \nonumber
&= \sum_{t=0}^{\infty}  \gamma^t \mathop{\mathbb{E}}_{s_t \sim p_{\tilde{\theta}}(s_t), a_t \sim \pi_{\tilde{\theta}}(a_t|s_t)} \left [ \frac{\pi_{\theta}(a_t|s_t)}{\pi_{\tilde{\theta}}(a_t|s_t)} A_{\tilde{\theta} }(s_t, a_t) \right ] \\
&= \mathop{\mathbb{E}}_{\tau \sim p_{\tilde{\theta}}(\tau)} \left [ \sum_{t=0}^{\infty} \gamma^t \frac{\pi_{\theta}(a_{t}|s_{t})}{\pi_{\tilde{\theta}}(a_{t}|s_{t})} A_{\tilde{\theta} }(s_t, a_t) \right ]
\end{align}

\

\subsection{Proposition 4} 
\label{ap:ap:prop4sec}
\paragraph{Proposition 4.} \label{ap:propos4}
 (Hindsight Expected Return). For the original goal $g$ and an hindsight goal $g'$, the object function of HTRPO $ L_{\tilde{\theta}}(\theta) $ is given by:
\begin{align}
L_{\tilde{\theta}}(\theta)
&= \mathop{\mathbb{E}}_{g'} \left [  \mathop{\mathbb{E}}_{\tau \sim p_{\theta}(\tau | g)} \left[ \sum_{t=0}^{\infty} \prod_{k=0}^{t} \frac{\pi_{\tilde{\theta}}(a_k | s_k, g')}{\pi_{\tilde{\theta}}(a_k | s_k, g)} \gamma^t  R_t \right ] \right ]
\end{align}
in which, $\tau =  s_0, a_0, s_1, a_1, ... , s_t, a_t$, and $R_t=\frac{\pi_\theta(a_t|s_t, g')}{\pi_{\tilde{\theta}}(a_t|s_t, g')}  A_{\tilde{\theta}}(s_t, a_t, g')$.

\proof Starting from:
\begin{align}
L_{\tilde{\theta}}(\theta) &= \mathop{\mathbb{E}}_{g, \tau} \left [ \sum_{t=0}^{\infty} \gamma^t \frac{\pi_{\theta}(a_{t}|s_{t}, g)}{\pi_{\tilde{\theta}}(a_{t}|s_{t}, g)} A_{\tilde{\theta} }(s_t, a_t, g) \right ] \label{ap:eq2.1}
\end{align}
for every time step $t$ in the expectation, denote
\begin{align}
L_{\tilde{\theta}}(\theta, t) &= \mathop{\mathbb{E}}_{g} \left [ \mathop{\mathbb{E}}_{\tau \sim p_{\tilde{\theta}}(\tau | g)} \left [ \gamma^t \frac{\pi_{\theta}(a_{t}|s_{t}, g)}{\pi_{\tilde{\theta}}(a_{t}|s_{t}, g)} A_{\tilde{\theta} }(s_t, a_t, g) \right ] \right ] ,
\end{align}
so that
\begin{align}
L_{\tilde{\theta}}(\theta)
&= \sum_{t=0}^{\infty}L_{\tilde{\theta}}(\theta, t).
\end{align}
Split every trajectory $\tau$ into $\tau_1$ and $\tau_2$ where $\tau_1 = s_0, a_0, s_1, a_1, ... , s_t, a_t$ and $\tau_2 = s_{t+1}, a_{t+1}, ...$, then
\begin{align}
&L_{\tilde{\theta}}(\theta, t) \nonumber \\
&= \mathop{\mathbb{E}}_{g} \left [ \mathop{\mathbb{E}}_{\tau_1 \sim p_{\tilde{\theta}}(\tau_1 | g)} \left [ \mathop{\mathbb{E}}_{\tau_2 \sim p_{\tilde{\theta}}(\tau_2 | \tau_1, g)} \left [ \gamma^t \frac{\pi_{\theta}(a_{t}|s_{t}, g)}{\pi_{\tilde{\theta}}(a_{t}|s_{t}, g)} A_{\tilde{\theta} }(s_t, a_t, g) \right ] \right ] \right ]
\end{align}
For that $\gamma^t \frac{\pi_{\theta}(a_{t}|s_{t}, g)}{\pi_{\tilde{\theta}}(a_{t}|s_{t}, g)} A_{\tilde{\theta} }(s_t, a_t, g)$ is independent from $\tau_2$ conditioned on $\tau_1$,
\begin{align} \nonumber
L_{\tilde{\theta}}(\theta, t) = \mathop{\mathbb{E}}_{g} \left [ \mathop{\mathbb{E}}_{\tau_1 \sim p_{\tilde{\theta}}(\tau_1 | g)} \left [ \gamma^t \frac{\pi_{\theta}(a_{t}|s_{t}, g)}{\pi_{\tilde{\theta}}(a_{t}|s_{t}, g)} A_{\tilde{\theta} }(s_t, a_t, g) \right ] \ \right ] 
\end{align}
Thus,
\begin{align}
L_{\tilde{\theta}}(\theta) & = \sum_{t=0}^{\infty} \mathop{\mathbb{E}}_{g} \left [ \mathop{\mathbb{E}}_{\tau_1 \sim p_{\tilde{\theta}}(\tau_1 | g)} \left [ \gamma^t \frac{\pi_{\theta}(a_{t}|s_{t}, g)}{\pi_{\tilde{\theta}}(a_{t}|s_{t}, g)} A_{\tilde{\theta} }(s_t, a_t, g) \right ] \ \right ]
\end{align}
Following the techniques of importance sampling, the objective function can be rewritten in the form of new goal $g'$ :
\begin{align}
L_{\tilde{\theta}}(\theta) & = \sum_{t=0}^{\infty} \mathop{\mathbb{E}}_{g'} \left [ \mathop{\mathbb{E}}_{\tau_1 \sim p_{\tilde{\theta}}(\tau_1 | g)} \left [ \frac{p_{\tilde{\theta}}(s_{0:t}, a_{0:t} | g')}{p_{\tilde{\theta}}(s_{0:t}, a_{0:t} | g)} \gamma^t R_t \right ] \ \right ] .
\end{align}
where $R_t=\frac{\pi_{\theta}(a_{t}|s_{t}, g')}{\pi_{\tilde{\theta}}(a_{t}|s_{t}, g')} A_{\tilde{\theta} }(s_t, a_t, g')$. Furthermore, given that
\begin{align}
p(s_{0:t}, a_{0:t} | g) = p(s_0)p(a_t | s_t, g) \prod_{k=0}^{t-1} p(a_k | s_k, g) p(s_{k+1} | s_k, a_k) , \label{ap:eqb.1}
\end{align}
after expanding the objective function and cancelling terms,
\begin{align} \nonumber
L_{\tilde{\theta}}(\theta)
&= \sum_{t=0}^{\infty} \mathop{\mathbb{E}}_{g'} \left [ \mathop{\mathbb{E}}_{\tau_1 \sim p_{\tilde{\theta}}(\tau_1 | g)} \left[ \prod_{k=0}^{t} \frac{\pi_{\tilde{\theta}}(a_k | s_k, g')}{\pi_{\tilde{\theta}}(a_k | s_k, g)} \gamma^t  R_t \right ] \right ] \\
\label{ap:eqb.2}
\end{align}

\
\subsection{Proposition 5} \label{ap:ap7}
\paragraph{Proposition 5.} (Quadratic KL divergence Estimation). For the original goal $g$ and a hindsight goal $g'$, the constraint between policy $\pi_{\tilde{\theta}}(a|s)$ and policy $\pi_{\theta}(a|s)$ is given by:
\begin{align}
\mathop{\mathbb{E}}_{g'} \left [ \mathop{\mathbb{E}}_{\tau \sim p_{\theta}(\tau | g)} \left [  \sum_{t = 0} ^\infty \prod_{k=1}^{t} \frac{\pi_{\tilde{\theta}}(a_k | s_k, g')}{\pi_{\tilde{\theta}}(a_k | s_k, g)} \gamma^t  K_t\right] \right ] \leq \epsilon'
\end{align}
in which $\epsilon' = \frac{\epsilon}{1 - \gamma}$, and $K_t=\frac{1}{2} (\log \pi_{\tilde{\theta}}(a_t|s_t, g') - \log \pi_{\theta}(a_t|s_t,g'))^2$.

\proof
Starting from the QKL-TRPO constraint:
\begin{align}
\mathop{\mathbb{E}}_{s, a \sim \rho_{\tilde{\theta}}(s,a)} \left[ \frac{1}{2} (\log \pi_{\tilde{\theta}}(a|s) - \log \pi_{\theta}(a|s))^2 \right] \leq \epsilon
\end{align}
Let $\tilde{K}_t=\frac{1}{2} (\log \pi_{\tilde{\theta}}(a|s) - \log \pi_{\theta}(a|s))^2$ and multiply the constraint function by a constant $\frac{1}{1 - \gamma}$,
\begin{align}
\frac{1}{1 - \gamma} \mathop{\mathbb{E}}_{s, a \sim \rho_{\tilde{\theta}}(s,a)} \left[\tilde{K}_t \right] \leq \frac{\epsilon}{1 - \gamma}
\end{align}
Denote the constraint function as $f_{\tilde{\theta}}(\theta)$,
\begin{align} \nonumber
f_{\tilde{\theta}}(\theta)
&= \frac{1}{1 - \gamma}\sum_{s \in \mathcal{S}} \frac{\sum_{t = 0}^{\infty} \gamma^t P(s_t = s)}{\frac{1}{1-\gamma}} \mathop{\mathbb{E}}_{ a \sim \pi_{\tilde{\theta}}(a|s)} \left[ \tilde{K}_t \right] \\ \nonumber
&= \sum_{t = 0}^{\infty} \gamma^t \mathop{\mathbb{E}}_{ s_t \sim p_{\tilde{\theta}}(s_t), a_t \sim \pi_{\tilde{\theta}}(a_t|s_t)} \left[ \tilde{K}_t \right] \\
&= \mathop{\mathbb{E}}_{\tau \sim p_{\tilde{\theta}}(\tau)} \left[\sum_{t = 0} ^ \infty \gamma^t \tilde{K}_t \right]
\end{align}
To write the constraint function in goal-conditioned form, let $\tilde{K}_{t,g}=\frac{1}{2} (\log \pi_{\tilde{\theta}}(a_t|s_t, g) - \log \pi_{\theta}(a_t|s_t,g))^2$
\begin{align}
f_{\tilde{\theta}}(\theta) = \mathop{\mathbb{E}}_{g} \mathop{\mathbb{E}}_{\tau \sim p_{\tilde{\theta}}(\tau |g)} \left[ \sum_{t = 0} ^\infty \gamma^t \tilde{K}_{t,g} \right]
\end{align}
In a similar way with the proof for Proposition \ref{ap:propos4}, denote every time step of $f_{\tilde{\theta}}(\theta)$ as $f_{\tilde{\theta}}(\theta, t)$, in other words,
\begin{align}
f_{\tilde{\theta}}(\theta) = \sum_{t = 0} ^ \infty f_{\tilde{\theta}}(\theta, t)
\end{align}
Let trajectory $\tau_1 = s_0, a_0, s_1, a_1, ... , s_t, a_t$ and $\tau_2 = s_{t+1}, a_{t+1}, ...$,
\begin{align}
f_{\tilde{\theta}}(\theta, t) = \mathop{\mathbb{E}}_{g} \left [ \mathop{\mathbb{E}}_{\tau_1 \sim p_{\tilde{\theta}}(\tau_1 | g)}  \left [ \gamma^t \tilde{K}_{t,g} \right] \right ]
\end{align}
for that $\tilde{K}_{t,g}$ is independent from $\tau_2$ conditioned on $\tau_1$.
Furthermore, by importance sampling, for a new goal $g'$, the constraint can be converted to the following form
\begin{align}
f_{\tilde{\theta}}(\theta) = \mathop{\mathbb{E}}_{g'} \left [ \mathop{\mathbb{E}}_{\tau \sim p_{\theta}(\tau | g)} \left [ \sum_{t = 0} ^\infty   \frac{p_{\tilde{\theta}}(s_{1:t}, a_{1:t} | g')}{p_{\tilde{\theta}}(s_{1:t}, a_{1:t} | g)} \gamma^t K_t \right] \right ].
\end{align}
in which $\tau = s_1, a_1, s_2, a_2, ... , s_t, a_t$, and $K_t = \frac{1}{2} (\log \pi_{\tilde{\theta}}(a_t|s_t, g') - \log \pi_{\theta}(a_t|s_t,g'))^2$.
Denote $\epsilon' = \frac{\epsilon}{1 - \gamma}$. Based on eq.\ref{ap:eqb.1}, by expanding and canceling terms, the constraint condition can be written as
\begin{align}
\mathop{\mathbb{E}}_{g'} \left [ \mathop{\mathbb{E}}_{\tau \sim p_{\theta}(\tau | g)} \left [\sum_{t = 0} ^\infty \prod_{k=0}^{t} \frac{\pi_{\tilde{\theta}}(a_k | s_k, g')}{\pi_{\tilde{\theta}}(a_k | s_k, g)} \gamma^t K_t \right] \right ] \leq \epsilon'.
\end{align}

\section{Solving Process for HTRPO} \label{ap:ap8}

\subsection{HTRPO Estimators}

Based on the final form of HTRPO optimization problem, this section completes the feasibility of this algorithm with estimators for the objective function and the KL-divergence constraint.

Given a dataset of trajectories and goals $\mathcal{D} = \{ \bm{\tau}^{(i)}, g^{(i)} \}^{N_{\tau}}_{i=1}$, each trajectory $\bm{\tau}^{(i)}$ is obtained from interacting with the environment under a goal $g^{(i)}$. In order to generate hindsight experience, we also need to sample a set of hindsight goals $\mathcal{G} = \{ g'^{(i)} \} ^{N_g}_{i = 1}$ . The Monte Carlo estimation of HTRPO optimization problem with dataset $\mathcal{D}$ can be derived as follows:
\begin{align}
\max_{\theta} \ \frac{1}{\lambda}\sum_{g',\tau,t} \left [ \gamma^t R_t \prod_{k=0}^{t} \frac{\pi_{\tilde{\theta}}(a_k | s_k, g')}{\pi_{\tilde{\theta}}(a_k | s_k, g)}  \right ]
\end{align}
\begin{align}
s.t. \ \frac{1}{\lambda} \sum_{g',\tau,t} \left [ \gamma^t K_t \prod_{k=0}^{t}  \frac{\pi_{\tilde{\theta}}(a_k | s_k, g')}{\pi_{\tilde{\theta}}(a_k | s_k, g)} \right] \leq \epsilon',
\end{align}
in which $\lambda = N_\tau \cdot N_g$, $R_t= \frac{\pi_\theta(a_k | s_k, g')}{\pi_{\tilde{\theta}}(a_k | s_k, g')}  A_{\tilde{\theta}}(a_k, s_k, g')$, and $K_t=\frac{1}{2}(\log \pi_{\tilde{\theta}}(a_k | s_k,  g') - \log \pi_{\theta}(a_k | s_k, g'))^2$. The hindsight goal $g'$ is supposed to follow original goal distribution to conduct the unbiased optimization and guide the agent to reach original goals.

However, during the hindsight learning process, the algorithm is to guide the agent to achieve the hindsight goals and then generalize to the original goals. Therefore, in experiments, we follow the Hindsight Goal Filtering method introduced in Algorithm \ref{ap:HGF} to select conductive hindsight goals. As a result, the goals of training data actually follow the distribution of hindsight goals instead of original goals, and the objective and KL expectation will be estimated w.r.t. the hindsight goal distribution. Such mechanism serves as a mutual approach for all hindsight methods \cite{andrychowicz2017hindsight,rauber2018hindsight,plappert2018multi}, which can be seen as a merit.

However, as discussed in \cite{rauber2018hindsight}, this kind of estimation may result in excessive variance, which leads to an unstable learning curve. In order to avoid instability, we adopt the technique of weighted importance sampling introduced in \cite{bishop2016pattern} and further convert the optimization problem to the following form:
\begin{align}
\max_{\theta} \ \frac{1}{\lambda} \sum_{g',\tau,t}  \left [ \gamma^t  R_t \frac{\overset{t} {\underset{k=0}{\prod}}   \frac{\pi_{\tilde{\theta}}(a_k | s_k, g')}{\pi_{\tilde{\theta}}(a_k | s_k, g)}}{ {\underset{\tau}{\sum}}  \overset{t} {\underset{k=0}{\prod}} \frac{\pi_{\tilde{\theta}}(a_k | s_k, g')}{\pi_{\tilde{\theta}}(a_k | s_k, g)}}  \right ] \label{ap:eq4.1}
\end{align}
\begin{align}
s.t. \ \frac{1}{\lambda} \sum_{g',\tau,t} \left [\gamma^t K_t \frac{\overset{t} {\underset{k=0}{\prod}}   \frac{\pi_{\tilde{\theta}}(a_k | s_k, g')}{\pi_{\tilde{\theta}}(a_k | s_k, g)}}{ {\underset{\tau}{\sum}}  \overset{t} {\underset{k=0}{\prod}} \frac{\pi_{\tilde{\theta}}(a_k | s_k, g')}{\pi_{\tilde{\theta}}(a_k | s_k, g)}} \right] \leq \epsilon'. \label{ap:eq4.2}
\end{align}
We provide an explicit solution method for the optimization problem above in Appendix \ref{ap:ap8.2}.

While introducing weighted importance sampling may cause a certain level of bias which is identical to that of HPG \cite{rauber2018hindsight}, the bias is to decrease in inverse ratio with regard to the increase of the data theoretically \cite{powell1966weighted}. Given the limited resources, we need to tradeoff between reducing bias and enlarging batch size. By picking a appropriate batch size, the improvement of weighted importance sampling is well demonstrated in the experiments.

\
\subsection{Solution Method for HTRPO} \label{ap:ap8.2}
For the HTRPO optimization problem, briefly denote the optimization problem in expression \ref{ap:eq4.1} and \ref{ap:eq4.2} as:
\begin{equation} \nonumber
\max_{\theta} \ f(\theta)
\end{equation}
\begin{equation}
s.t. \ g(\theta) \leq \epsilon'.
\end{equation}
For any policy parameter $\theta$ in the neighborhood of the parameter $\tilde{\theta}$, approximate the optimization problem with linear objective function and quadratic constraint:
\begin{equation} \nonumber
\max_{\theta} \ f(\tilde{\theta}) + \nabla_\theta f(\tilde{\theta})(\theta - \tilde{\theta})
\end{equation}
\begin{equation}
s.t. \ g(\tilde{\theta}) + \nabla_\theta g(\tilde{\theta})(\theta - \tilde{\theta}) + \frac{1}{2}(\theta - \tilde{\theta})^T \nabla^2_\theta g(\tilde{\theta})(\theta - \tilde{\theta}) \leq \epsilon'.
\end{equation}
Noticeably, $g(\tilde{\theta}) = 0$ and $\nabla_\theta g(\tilde{\theta}) = 0$, which further simplifies the optimization problem to the following form:
\begin{equation} \nonumber
\max_{\theta} \ f(\tilde{\theta}) + \nabla_\theta f(\tilde{\theta})(\theta - \tilde{\theta})
\end{equation}
\begin{equation}
s.t. \ \frac{1}{2}(\theta - \tilde{\theta})^T \nabla^2_\theta g(\tilde{\theta})(\theta - \tilde{\theta}) \leq \epsilon'.
\end{equation}
Given a convex optimzation problem with a linear objective function under a quadratic constraint, many well-practiced approaches can be taken to solve the problem analytically, among which we adopt the Karush-Kuhn-Tucker(KKT) conditions \cite{boyd2004convex}. For a Lagrangian multiplier $\lambda$,
\begin{align}  \nonumber
\frac{1}{2}(\theta - \tilde{\theta})^T \nabla^2_\theta g(\tilde{\theta})(\theta - \tilde{\theta}) - \epsilon' \leq 0 	\nonumber \\
\lambda \geq 0	\nonumber  \\
\lambda [\frac{1}{2}(\theta - \tilde{\theta})^T \nabla^2_\theta g(\tilde{\theta})(\theta - \tilde{\theta}) - \epsilon'] = 0  \nonumber \label{ap:eqd.1} \\
-\nabla_\theta f(\tilde{\theta}) + \lambda \nabla^2_\theta g(\tilde{\theta}) (\theta - \tilde{\theta}) = 0 	
\end{align}
Expressions in \ref{ap:eqd.1} form the KKT conditions of the optimization problem. Solving the KKT conditions,
\begin{equation}
\theta = \tilde{\theta} + \sqrt{\frac{2 \epsilon'}{\nabla_\theta f(\tilde{\theta})^T [\nabla^2_\theta g(\tilde{\theta})]^{-1} \nabla_\theta f(\tilde{\theta})}} [\nabla^2_\theta g(\tilde{\theta})]^{-1} \nabla_\theta f(\tilde{\theta})
\end{equation}
The policies, however, in this paper are in the form of a neural network, which makes it extremely time-comsuming to compute the Hessian matrix. Thus, we compute $[\nabla^2_\theta g(\tilde{\theta})]^{-1} \nabla_\theta f(\tilde{\theta})$ with conjugate gradient algorithm by solving the following equation:
\begin{equation}
[\nabla^2_\theta g(\tilde{\theta})] x = \nabla_\theta f(\tilde{\theta}),
\end{equation}
in which $[\nabla^2_\theta g(\tilde{\theta})] x$ can be practically calculated through the following expansion:
\begin{equation}
[\nabla^2_\theta g(\tilde{\theta})] x = \nabla_\theta [ (\nabla_\theta g(\tilde{\theta}))^T x].
\end{equation}

\section{Algorithms} \label{ap:ap_alg}

\subsection{Hindsight Goal Filtering}\label {ap:HGF}

The algorithm of hindsight goal filtering is shown in Algorithm \ref{ap:alg-hgf}.

\begin{algorithm} 
\caption{Hindsight Goal Filtering}
\begin{algorithmic}[1]
\label{ap:alg-hgf}
\STATE {\bfseries Input:} \\
 Set of achieved goals $\mathcal{G}_a$\\
 Set of original goals $\mathcal{G}_o$\\
 Hindsight goal number $N$

\STATE {\bfseries Output:} \\ Set of hindsight goals $\mathcal{G}$
\STATE Valid achieved goal set $\mathcal{G}_v = \mathcal{G}_a \cap \mathcal{G}_o$
 \IF{$\mathcal{G}_v = \varnothing$}
 \FOR{$i = 1$ to $\min{(N, |\mathcal{G}_a|)}$}
 \STATE $g^* = \mathop{\arg\min}_{g \in \mathcal{G}_a - \mathcal{G}}{d(g, \mathcal{G}_o)}$
 \STATE $\mathcal{G} = \mathcal{G} \cup \{g^*\}$
 \ENDFOR
 \ELSE
 \STATE Randomly sample $g_0 \in \mathcal{G}_v$, $\mathcal{G} = \{g_0\}$
 \FOR {$i = 1$ to $\min{(N-1, |\mathcal{G}_v - \mathcal{G}|)}$}
 \STATE $g^* = \arg\max_{g \in \mathcal{G}_v - \mathcal{G}}{\min_{g' \in \mathcal{G}}{d(g, g')}}$
 \STATE $\mathcal{G} = \mathcal{G} \cup \{g^*\}$
 \ENDFOR
 \ENDIF \\
 \STATE {\bfseries Return:} $\mathcal{G}$
 \end{algorithmic}
\end{algorithm}

\subsection{Hindsight Trust Region Policy Optimization}\label {ap:HTRPO}

The complete algorithm of HTRPO can be found in Algorithm \ref{ap:alg-htrpo}. It iteratively solves a local optimization problem to find the optimal solution until the reward converges.

\begin{algorithm}[h]

\caption{Hindsight Trust Region Policy Optimization}
\begin{algorithmic}[1]
\label{ap:alg-htrpo}
\STATE {\bfseries Input:} \\
 Training batch size $batchsize$\\
 Max number of training time steps $T_{max}$ \\
 Policy $\theta$ \\
 Q-function $\phi$ \\
 Data Buffer $B_{origin}$ with its max size equal to $batchsize$ 
\STATE {\bfseries Output:} \\ Optimized Policy $\theta^{*}$
\FOR{$iteration=1$ to $T_{max} / batchsize$}
 \WHILE{$B_{origin}$ is not full}
 \STATE Sample a trajectory $\tau=\{(s_t, a_t, r_t, s_{t+1}, g,$ $\pi_{\theta}(a_t|s_t,g))\}_{t=1}^{T}$ using current policy $\theta$;
 \STATE $B_{origin} = B_{origin}\cap \tau$;
 \ENDWHILE  
 \STATE Sample hindsight goals $\mathcal{G} = \{ g'^{(i)} \} ^{N_g}_{i = 1}$ from achieved goal set $\mathcal{G}_a$ derived from $B_{origin}$ according to HGF (Appendix \ref{ap:ap_alg}, \textbf{Algorithm \ref{ap:HGF}});
 \STATE $B_{train}=\varnothing$;
 \FOR {$g'^{(i)}$ in $\mathcal{G}$}
 \FOR {$\tau$ in $B_{origin}$}
 \FOR {$t=0$ to T}
 \STATE Compute $\pi_{\theta}(a_t|s_t,g'^{(i)})$;
 \STATE Modify reward $r_t|g \rightarrow r_t|g'^{(i)}$;
 \ENDFOR
 \STATE $\tau|g'^{(i)} = \{(s_t, a_t, r_t|g'^{(i)}, s_{t+1}, g, \pi_{\theta}(a_t|s_t,g),$ $ \pi_{\theta}(a_t|s_t,g'^{(i)}))\}_{t=1}^{T}$;
 \STATE $B_{train} = B_{train}\cap\tau|g'^{(i)}$;
 \ENDFOR
 \ENDFOR		
 \STATE Use $B_{train}$ to optimize policy $\theta$ with objective (eq. \ref{ap:eq4.1}) and constraint (eq. \ref{ap:eq4.2}) following Section \ref{ap:ap8.2};
 \STATE $B_{origin}=\varnothing$;
 \ENDFOR
 \STATE {\bfseries Return:} optimized policy $\theta^{*}$;
 \end{algorithmic}[1]
\end{algorithm}

\section{Environments} 

\subsection{Environmental Settings} \label{ap:ap9}

In simple benchmarks, the network architecture is of two hidden layers, each with 64 units; in Ms. Pan-Man, the network is 3 convolutional layers (8 filters, stride 4; 16 filters, stride 2; 16 filters, stride 2) concatenating 3 32-unit fully-connected layers; in robot environment, the network contains three 256-unit hidden layers.

 \begin{table*}[!t]
 \caption{Hyperparameters of Discrete Environments}
 \label{ap:hpd}
 \begin{center}
 \begin{tabular}{lcccc}
 \toprule
 & Ms. Pac-Man & Fetch Reach & Fetch Push & Fetch Slide\\
\midrule
 training time steps & $2 \times 10^6$ & $2 \times 10^6$ & $2 \times 10^6$ & $2 \times 10^6$\\
 batch size &300 &1600 & 1600 & 1600\\
 cg damping &1e-3 &1e-3 &1e-3 & 1e-3\\
 reward decay &0.98 &0.98 &0.98 & 0.98\\
 max KL step  &2e-5 &2e-5 &2e-5 & 2e-5\\
 sampled goal number & 16 & 100 & 100 & 100\\
 critic optimizer &Adam &Adam &Adam &Adam\\
 critic learning rate & 5e-5 & 5e-4 &5e-4 &5e-4\\
 critic updates per iteration &10 & 20 &20 &20\\
\bottomrule
 \end{tabular}
 \end{center}
\end{table*}

 \begin{table*}[!t]
 \caption{Hyperparameters of Continuous Environments}
 \label{ap:hpc}
 \begin{center}
 \begin{tabular}{lcccc}
\toprule
 & Fetch Reach & Fetch Push & Fetch Slide  & Fetch PickAndPlace \\
\midrule
 training time steps & $2 \times 10^6$ & $2 \times 10^6$ & $2 \times 10^6$ &  $5 \times 10^6$  \\
 batch size & 3200 & 3200 & 3200 & 3200  \\
 cg damping & 1e-3 & 1e-3 & 1e-3 & 1e-3 \\
 reward decay & 0.98 & 0.98 &0.98&0.98 \\
 max KL step & 2e-5 & 2e-5 & 2e-5 &2e-5  \\
 critic optimizer & Adam &Adam & Adam& Adam \\
 critic learning rate &5e-4 &5e-4 &5e-4&5e-4 \\
 critic updates per iteration & 20&20 &20&20  \\
 sampled goal number & 100 &100 &100&100 \\
\bottomrule
 \end{tabular}
 \end{center}
\end{table*}

\begin{savenotes}
\begin{table}[!t]
\centering
\begin{threeparttable}
\renewcommand\TPTminimum{3in}
 \caption{Control Mode for Different Robot Envs.}
 \label{ap:robot_env}
 \begin{tabular}{ll}
 \toprule
 Robot Envs & Control Mode \\
\midrule
 Discrete Fetch & Velocity Difference Control\\
 Continuous Fetch & End-Effector Velocity Control\\
\bottomrule
 \end{tabular}
\end{threeparttable}
\end{table}
\end{savenotes}

\textbf{k-Bit Flipping.} In each episode of this experiment, two arrays of length $k$ are generated. The first array is initialized with all 0's while the second one, usually regarded as the target array, is generated randomly. At each time step, the agent is able to flip one bit of the first array from 0 to 1 or from 1 to 0. Once the first array is exactly the same with the target array, the agent reaches the goal state and is then rewarded. The maximum number of time steps is $k$.


\textbf{Ms. Pac-Man.} Ms. Pac-Man is one of the ATARI 2600 games. We adopt its simulation in \cite{bellemare13arcade}. The agent is initialized in a $14 \times 19$ map with walls and moving enemies. Its goal is to reach a random position, which is represented by a 2-d vector, and the actions are to move up, down, left and right at each time step. The maximum number of time steps is 26. One episode also will terminate when the agent runs into an enemy. The state is represented by raw-image screenshots of the game interface, $210 \times 160$ in size with 3 channels. The image preprocessing follows \cite{rauber2018hindsight}, resulting in a $84\times 84$ image with 4 channels as the input of the convolutional neural network.

\textbf{Fetch.} Fetch environment contains a 7-DoF Fetch robotic arm with a two-fingered parallel gripper \cite{plappert2018multi}, which forms a 25-d state space. For discrete version of Fetch environments, we follow \cite{plappert2018multi} to attach an extra 4-d vector, which indicates the current velocity and gripper state of the end-effector, which results in a 29-d state space. The action space is 4-d, with 3-d end-effector velocities in \emph{x, y, z} direction and 1-d grasping indicator. In \textit{Fetch Reach} environment, a target position is randomly chosen and the gripper of Fetch robotic arm needs to be moved upon it. In \textit{Fetch Push}, the task for the robotic arm is to push a randomly placed block towards the goal state, anther randomly picked position. In \textit{Fetch Slide}, the robotic arm needs to exert a force on the block for it to slide towards a chosen goal at a certain distance. In \textit{Fetch PickAnd Place}, a cube is initialized randomly on the table as well as a target position. The target position is mostly up in the air. The robotic arm needs to firstly grasp the cube and then place it at the target position. All goals in the four tasks are represented by a 3-d Cartesian coordinate. For the discrete Fetch environment, detailed settings follow that in \cite{rauber2018hindsight}; for the continuous version, the configurations of legal actions and states follow that in \cite{plappert2018multi}. The maximum number of time steps is 50. The control modes are shown in Table \ref{ap:robot_env}.

\subsection{Hyperparameters} \label{ap:ap10}

Detailed hyperparameter settings can be found in Table \ref{ap:hpd} and \ref{ap:hpc}.

\section{Additional Experiments}

\subsection{Comparison between QKL-TRPO and TRPO}\label{ap:ap-qkltrpo}

\begin{figure*}[t]
 \vspace{-5pt} 
\center{\includegraphics[width=0.8\textwidth]{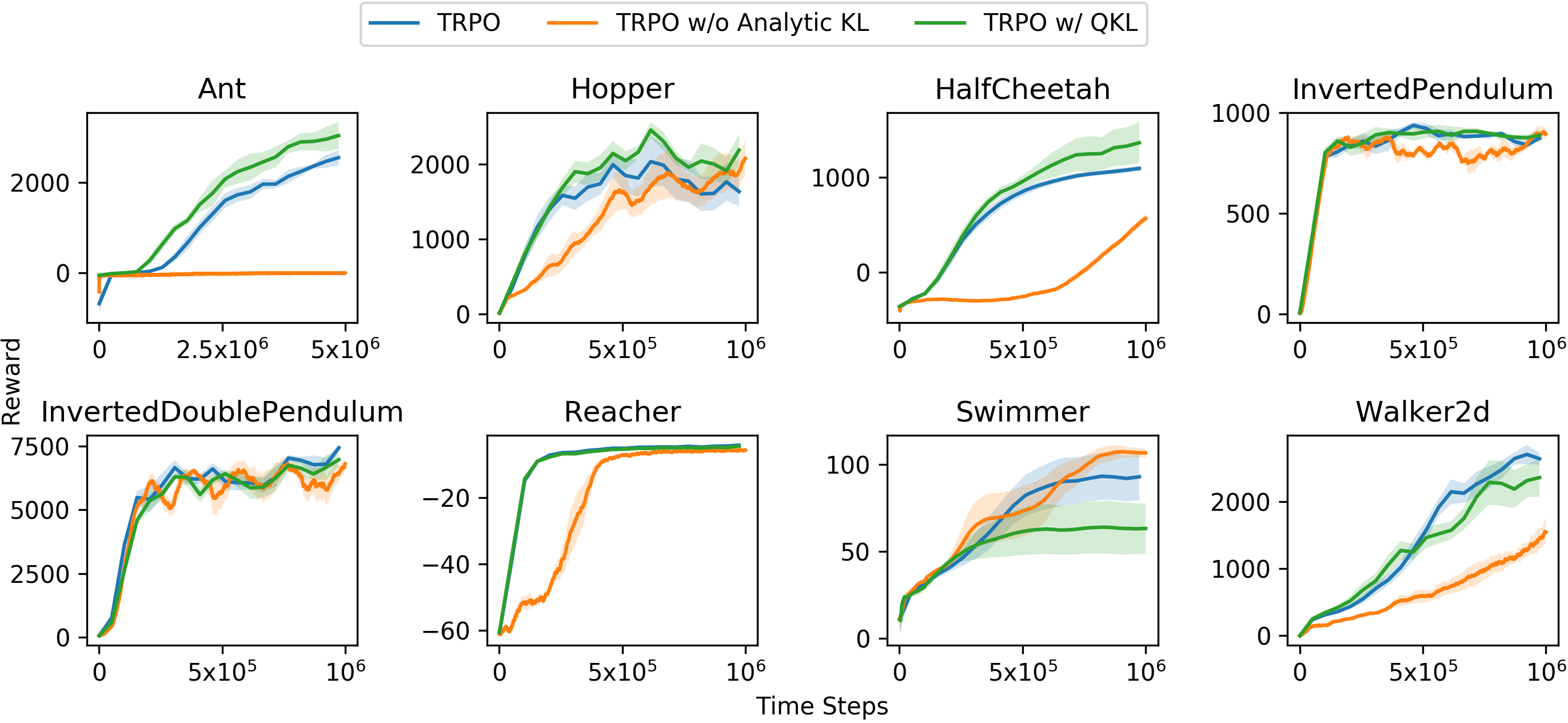}}
\caption{Reward of QKL-TRPO (TRPO w/ QKL), TRPO without analytical KL (TRPO w/o Analytic KL),  and the original TRPO (TRPO) in several MuJoCo benchmarks.}
\label{ap:fig4}
\end{figure*}

We test QKL-TRPO compared with the original TRPO in several MuJoCo benchmarks. We follow the same hyperparameters recommended in OpenAI Baseline. The only difference between these two versions of TRPOs is the method utilized to estimate KL divergence between old and new policies. From the results, we can see that QKL-TRPO performs better in (Ant, HalfCheetah, Hopper), comparable in (InvertedDoublePendulum, InvertedPendulum, Reacher), while worse in (Swimmer, Walker2d). Note that the KL divergence expectation is estimated by $\mathop{\mathbb{E}}_{s \sim \rho_{\tilde{\theta}}(s)} \left[ D_{KL}(\pi_{\tilde{\theta}}||\pi_{\theta}) \right]$ with the analytical form of KL divergence over all collected $s$ for the original TRPO while $\mathop{\mathbb{E}}_{s, a} \left[0.5( \log \pi_{\tilde{\theta}} - \log \pi_{\theta})^2 \right]$ over all $(s,a)$ pairs for QKL-TRPO. The latter $(s,a)$-based estimation method should have caused a much higher variance, resulting in lower performance (TRPO w/o Analytic KL in Figure \ref{ap:fig4}). However, thanks to QKL, QKL-TRPO achieves comparable performance with TRPO.

\subsection{Ablation Study} 

\begin{figure*}[t]
 \vspace{-5pt} 
 \center{\includegraphics[width=0.8\textwidth]{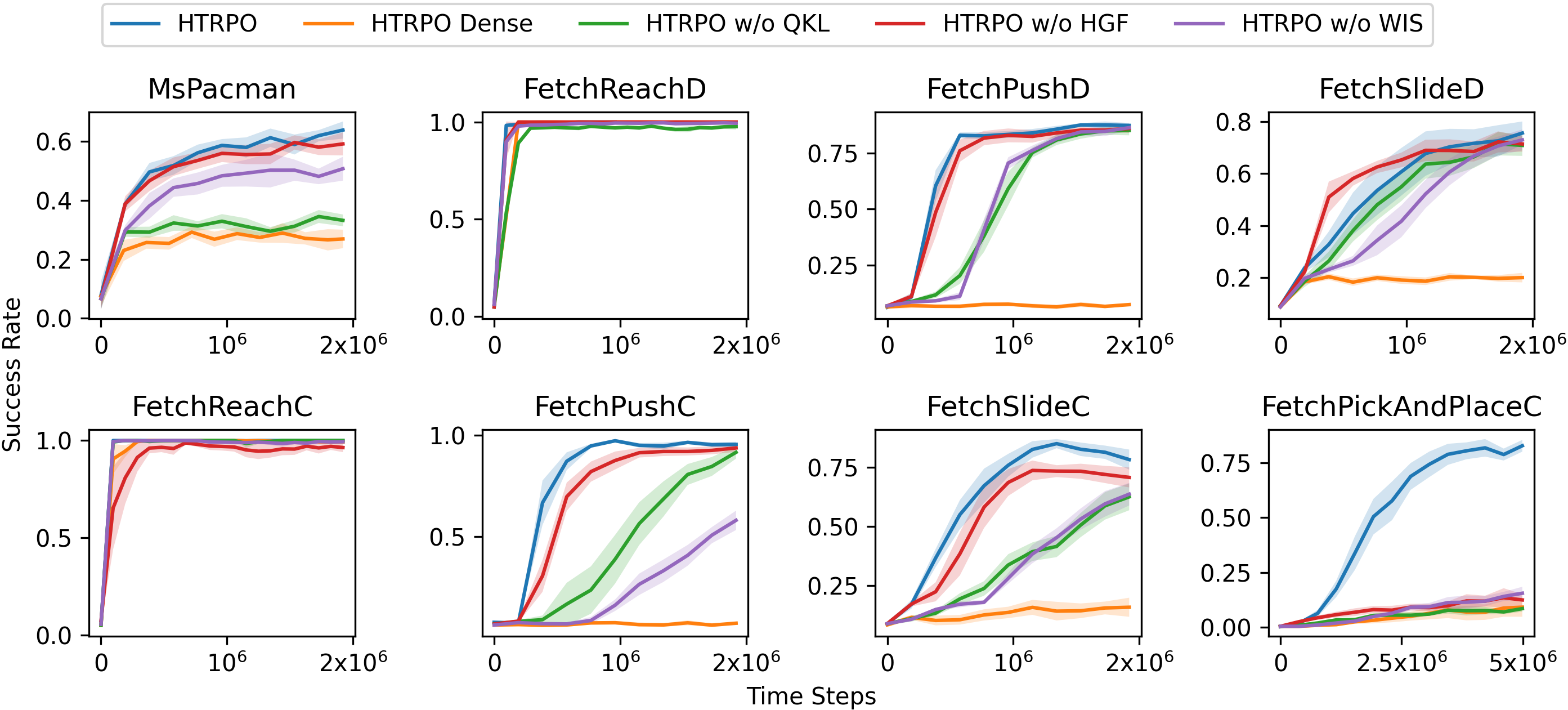}}       
 \caption{Ablation study of HTRPO.}  
  \vspace{-5pt} 
 \label{ap:fullabla} 
 \end{figure*}

We have conducted a series of ablation study to investigate the impacts of main components of HTRPO, Quadratic KL divergence estimation (QKL) and hindsight goal filtering (HGF), as well as the weighted importance sampling (WIS). Besides, we also provide a comprehensive comparison between dense-reward and sparse-reward HTRPO. All results are shown in Figure \ref{ap:fullabla}.
\
\label{ap:ap-abla}

\paragraph{Quadratic KL divergence Estimation:} QKL is a main component of HTRPO, which imposes a reliable trust region constraint for finding a local optimum. As is known, the vanilla KL estimation suffers from high variance and fails to give an accurate estimation, which leads to instability during learning. It accords with the results that ``HTRPO w/o QKL'' performs much worse than ``HTRPO" in both converging speed and final success rate.

\begin{figure}[t]
\vspace{-10pt}
\center{\includegraphics[width=0.5\textwidth]{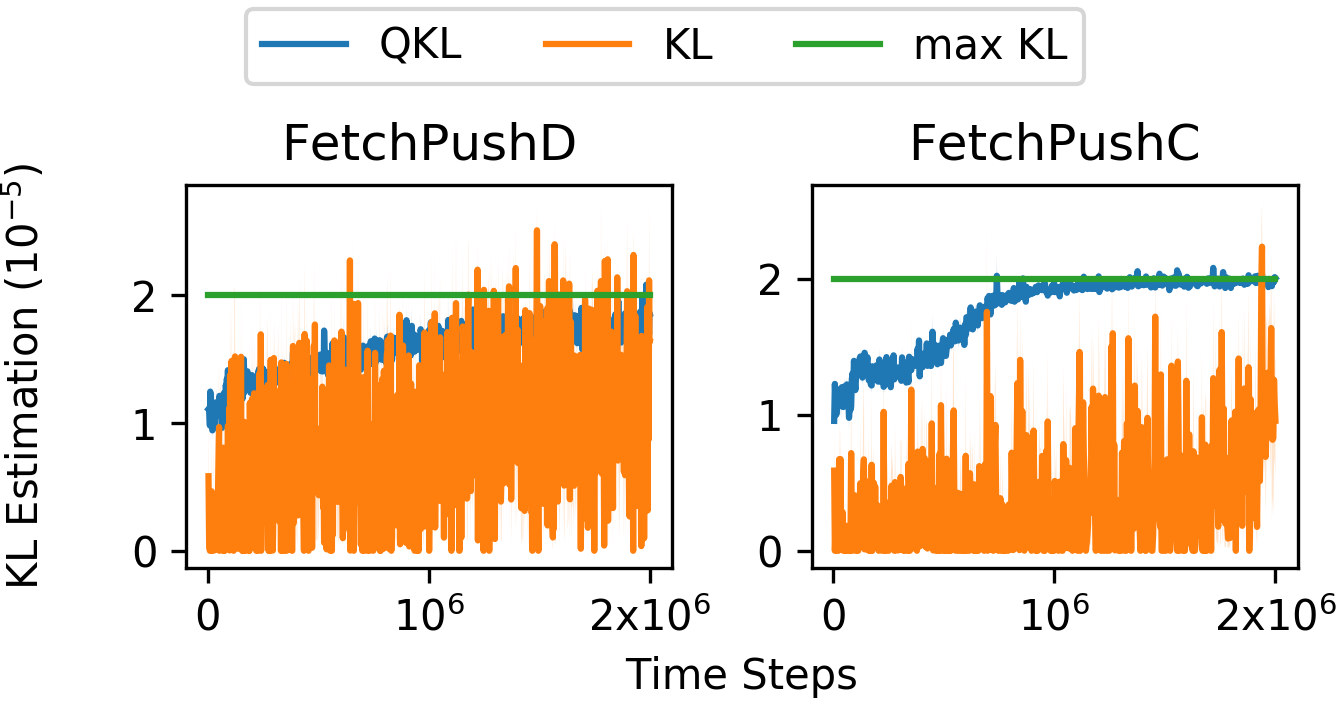}}
\vspace{-15pt}
\caption{Variance of different KL estimation methods.}
\label{ap:kl_est}
\end{figure}

To illustrate the advantage of QKL more intuitively, Figure \ref{ap:kl_est} provides the curves of KL estimation during the learning process for Discrete Fetch Push and Continuous Fetch PickAndPlace as examples. As known, the line search method in TRPO helps enhance learning stability. However, due to the high variance of vanilla KL estimation, when encountering an overestimated result, line search imposes a smaller update step. Such regulation underestimates the range of trust region and therefore, the vanilla KL estimation is randomly much lower than the expected KL(noted as max KL step). It introduces unstable update by finding the local optima in inappropriate trust regions. In comparison, the low variance of QKL ensures a comparatively accurate estimation and better accords with the max KL step. Still, in Figure \ref{ap:kl_est}, the KL curve of HTRPO is slightly below the max KL step, because we adopt a conjugate-gradient damping coefficient to enhance the training stability as in \cite{schulman2015trust}.

\paragraph{Hindsight Goal Filtering:} HGF scatters hindsight goals over the original goal space, which better guides the agent to successfully reach every possible original goal. To verify the contribution of HGF, we ablate this hindsight goal selecting method and provide the corresponding performance in Figure \ref{ap:fullabla}. As demonstrated, the algorithm would suffer a decrease in success rate more or less without HGF. Such degradation of performance becomes severe in complex tasks like Fetch PickAndPlace. Given that original goals mainly distribute in the air versus that most hindsight goals remain initially on the table, such discrepancy impedes the agent's progress of learning to reach the original goals. With the assistance of HGF, the selected hindsight goals cover the original goal space as much as possible, which, to a great extent, guarantees the successful reach to goals from the original goal space once the agent has learned to reach goals from the hindsight goal space. Hence, a good hindsight goal selection method brings significant benefit for hindsight data driven learning approaches.

\paragraph{Weighted Importance Sampling:} Weighted importance sampling is utilized for variance reduction of importance sampling \cite{bishop2016pattern}. In the experiment, we compare HTRPO with ``HTRPO without WIS''. Admittedly, we can see that the performance of ``HTRPO without WIS'' matches the full version of HTRPO in some simple tasks such as Continuous Fetch Reach. However, for more complex ones like Fetch Push and Fetch Slide, the variance is detrimentally larger than that in simple tasks. And we can see as the environment get more complex, the benefits of weighted importance sampling significantly improves the performance. In short, the performance of ``HTRPO without WIS'' has a severe degradation comparing to the full version of HTRPO.

To further demonstrate the sampling efficiency, we also follow a well-known method $\widehat{ESS}=1/\sum \overline{w}^2$, where $\overline{w}$ is the normalized importance weight, to estimate the effective sample sizes (ESS) of the importance weights by taking FetchPushC as an example. During learning, since we do have 3.2k time steps in each batch, by sampling 100 hindsight goals, there are about 320k importance-weighted samples. The corresponding ESS is 11.6k on average during the learning. It is also interesting that the ESS will decrease as the training progresses since the entropy of the policy will decrease.





\end{document}